\documentclass[letterpaper]{article} 
\usepackage{aaai23}  
\usepackage{times}  
\usepackage{helvet}  
\usepackage{courier}  
\usepackage[hyphens]{url}  
\usepackage{graphicx} 
\usepackage{natbib}  
\usepackage{caption} 
\usepackage{algorithm}
\usepackage{algorithmic}
\usepackage{subcaption}
\usepackage{newfloat}
\usepackage{listings}
\DeclareCaptionStyle{ruled}{labelfont=normalfont,labelsep=colon,strut=off} 
\floatstyle{ruled}
\newfloat{listing}{tb}{lst}{}
\floatname{listing}{Listing}
\title{CLIPVG: Text-Guided Image Manipulation Using Differentiable Vector Graphics}
\author{
    Yiren Song,\equalcontrib\textsuperscript{\rm 1,\rm 2}
    Xuning Shao,\equalcontrib\textsuperscript{\rm 2}
    Kang Chen,\textsuperscript{\rm 2}
    Weidong Zhang,\textsuperscript{\rm 2}
    Minzhe Li,\textsuperscript{\rm 1}
    Zhongliang Jing,\thanks{Corresponding author}\textsuperscript{\rm 1}
}
\affiliations {
    \textsuperscript{\rm 1}Shanghai Jiao Tong University, Shanghai, China,\\
    \textsuperscript{\rm 2}Netease Games AI Lab, Hangzhou, China,\\
    \{songyiren, liminzhe, zljing\}@sjtu.edu.cn,
    \{shaoxuning, ckn6763, zhangweidong02\}@corp.netease.com
}

\usepackage{bibentry}

\begin{document}

\maketitle

\begin{abstract}
Considerable progress has recently been made in leveraging CLIP (Contrastive Language-Image Pre-Training) models for text-guided image manipulation. However, all existing works rely on additional generative models to ensure the quality of results, because CLIP alone cannot provide enough guidance information for fine-scale pixel-level changes. In this paper, we introduce CLIPVG, a text-guided image manipulation framework using differentiable vector graphics,  which is also the first CLIP-based general image manipulation framework that does not require any additional generative models. We demonstrate that CLIPVG can not only achieve state-of-art performance in both semantic correctness and synthesis quality, but also is flexible enough to support various applications far beyond the capability of all existing methods.
\end{abstract}

\section{Introduction}

Large-scale vision-language pre-training models like CLIP (Contrastive Language-Image Pre-Training) significantly facilitate the task of text-guided image manipulation, whose goal is to automatically modify images based on given text prompts. In recent years, various studies \cite{styleclip, stylegannada, diffusionclip, dalle2, DiscoDiffusion} have been conducted on utilizing a pre-trained CLIP model for such purposes.

However, all existing CLIP-based works perform the manipulation on pixel-level, and thus share the same intrinsic limitation of raster image based methods, i.e., easily produce poor results. This is because CLIP cannot provide enough guidance for fine-scale pixel-level optimization since CLIP mainly focuses on high-level semantics of an image. As pointed out by \cite{stylegannada, glide}, the CLIP guided optimization process may be easily trapped in local optimum or impaired by adversarial solutions.

\begin{figure}[t!]
\centering
\includegraphics[width=\linewidth]{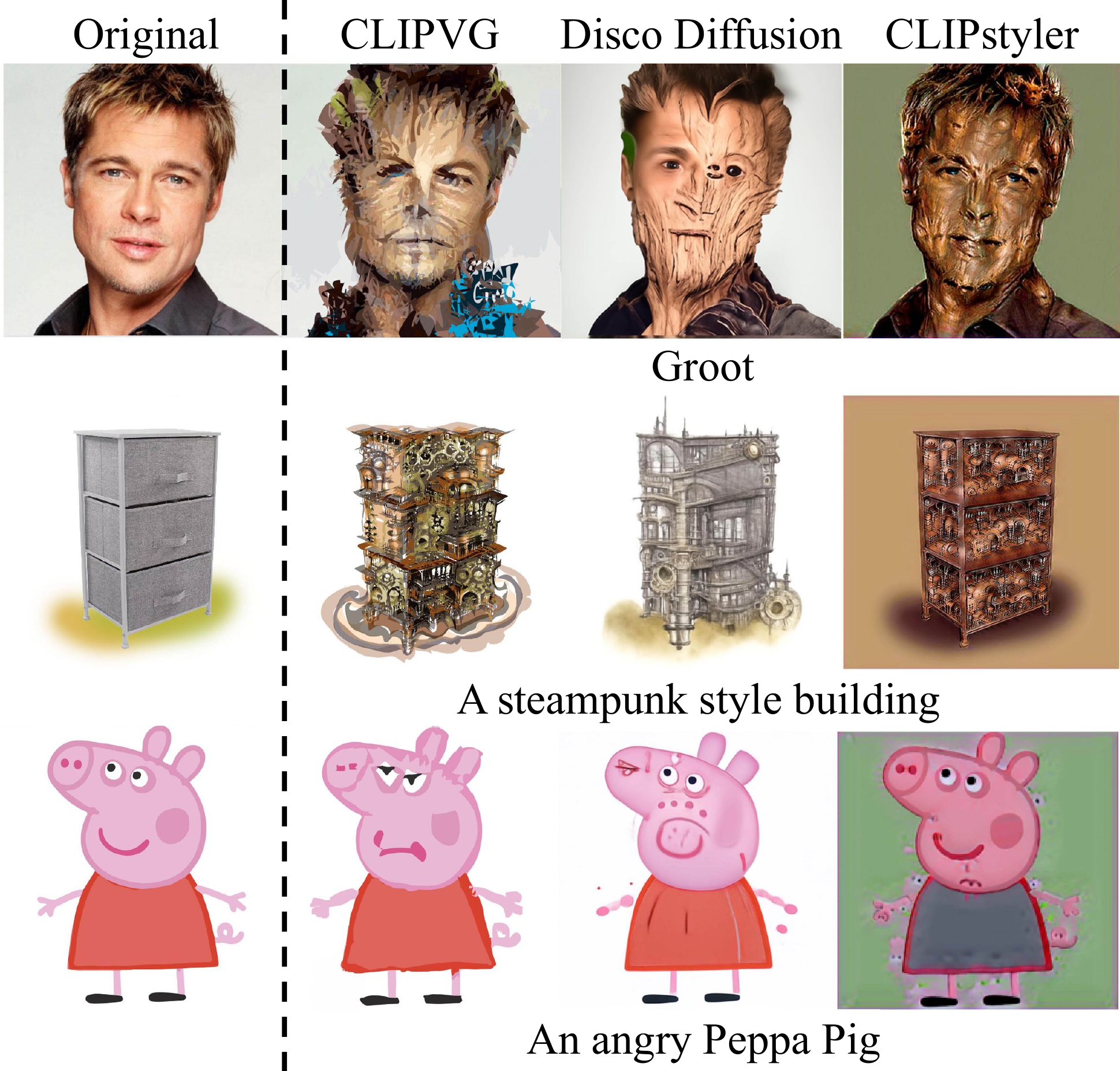}
\caption{Text-guided manipulation results of CLIPVG and two baselines, i.e., Disco Diffusion~\cite{DiscoDiffusion} and CLIPstyler~\cite{clipstyler}.} 
\label{fig:teaser}
\end{figure}

To mitigate such an issue, existing works typically incorporate additional generative models to ensure the synthesis quality \cite{styleclip, stylegannada, diffusionclip, DiscoDiffusion, dalle2}. These models not only consume extra resources to train, but limit the domain of the input images and text prompts. Currently, the only solution that does not rely on additional generative models is CLIPstyler~\cite{clipstyler}, which handles fine-scale features by additionally applying CLIP losses to a set of randomly sampled patches. However, this solution is only feasible for local texture style transfer, rather than general semantic manipulation (see Figure~\ref{fig:teaser}). 

In this paper, we tackle CLIP-based image manipulation from a new perspective. Specifically, we vectorize the input raster image into vector graphics using a robust multi-round vectorization strategy and leverage a differentiable 2D vector graphics rasterizer \cite{diffvg} to optimize the color and shape of each geometric element (i.e., stroke or filled curve) so that the CLIP loss between the text prompt(s) and the corresponding rasterized image can be minimized. The major difference between this new framework, which we call CLIPVG, and exiting works is that CLIPVG performs image manipulation in the domain of vector graphics. Since the vector graphical elements naturally function as some kind of regularization for local shape and color, the optimization process is significantly more stable when performed on the parameters of vector graphical elements (e.g., color, line width, control points, etc.) than on pixels. We surprisingly find that the effectiveness of such regularization in CLIP-based image manipulation is almost comparable to additionally incorporating a large-scale pre-trained generative model. As illustrated in Figure~\ref{fig:teaser}, CLIPVG better conforms to text semantics and produces fewer visual artifacts than Disco Diffusion~\cite{DiscoDiffusion} which relies on an additional diffusion model, i.e., \cite{ddpm}, trained on ImageNet~\cite{imagenet}.

Moreover, as a text-guided image manipulation framework, CLIPVG is much more flexible than existing frameworks. Firstly, CLIPVG inherits some advantages from vector graphics, i.e., it, by nature, is resolution-independent and allows separate manipulations of color and shape of each vector graphical element. Secondly, CLIPVG supports a wider range of applications (e.g., face attribute editing, character design, font design, re-colorization, etc.) since it does not bound to the domain of any specific pre-trained generative model. In other words, CLIPVG can fully unleash the capability of CLIP for image manipulation. It even allows users to assign different text prompts for different regions of the image at the same time.

The main contributions of this paper are:
\begin{itemize}
  \item We propose the first text-guided vector graphic manipulation framework which can achieve state-of-art performance without relying on any additional pre-trained models other than CLIP.
  \item We design a robust multi-round vectorization strategy which enables manipulation of raster images in the domain of vector graphics.
  \item We implement a flexible text-guided image manipulation system that supports a variety of controls far beyond the ability of all existing methods, and the source code of this system will be made publicly available.
\end{itemize}

\section{Related Works}
\textbf{Text-guided Image Manipulation.}
Pioneering studies \cite{dalle, cogview, manigan, talktoedit} model the relationship between text and image as a part of the image generation framework. Recently, the standalone CLIP models \cite{CLIP}, pre-trained on 400M text-image pairs, have shown a state-of-the-art performance in vision-language tasks. The latest methods \cite{styleclip, stylegannada, diffusionclip,VQGAN+CLIP, DiscoDiffusion, dalle2, anyface} typically use a CLIP model for parsing text-based guidance, and an additional generative model to constrain the output images. The generative model can be either trained on a specific category of images (domain-specific) or on a large database containing diverse image categories (domain-agnostic).

CLIP is often combined with a domain-specific StyleGAN \cite{Stylegan2} model for a human face, a cat, a church, etc. Given a text prompt, StyleCLIP \cite{styleclip} uses CLIP to find the corresponding manipulation direction in the latent space. StyleGAN-NADA \cite{stylegannada} adapts an existing StyleGAN model to a related domain defined by the prompt. It also proposes the directional CLIP loss to mitigate the mode collapse issue \cite{unrolledgan}. DiffusionCLIP replaces the GAN model with a diffusion model for image generation. These methods are generally robust, but their capabilities are tied to the domain of these pre-trained generative model.

There are also some domain-agnostic methods, most of which are developed for general-purpose image synthesis. The text-guided image synthesis methods \cite{dalle, dalle2, DiscoDiffusion, glide, imagen, VQGAN+CLIP, cogview, cogview2} typically support generating an image from a text prompt and a random latent code. To further enable image manipulation, some of these methods also provide an encoder to convert an input image to a corresponding latent code. For example, DALL·E-2 \cite{dalle2} and Disco Diffusion \cite{DiscoDiffusion} employ the diffusion process \cite{ddpm, ddim} as the encoder. These methods require massive data to train the general-purposed generative model. Moreover, additional upsampling models are often required to synthesize high resolution images \cite{dalle2, imagen, cogview2}.

Different from the above solutions, CLIPVG does not depend on any additional model other than CLIP. The output image is constrained by the vector graphic specific regularization rather than a generative or upsampling model.

Recently, a domain-agnostic method CLIPstyler \cite{clipstyler} is proposed to eliminate the dependency on the generative model. CLIPstyler applies the CLIP loss to a set of small randomly cropped patches to stabilize the optimization. The patch level CLIP loss helps to constrain the low level details of the image, and suppress the adversarial artifacts. However, since the text prompt is applied to each small patch, the use case is limited to the low level style transfer. In contrast, we relax the patch-wise constraints and support the general semantic manipulation.

Compared to all the raster image based solutions, our vector graphic based method also has other native benefits such as the infinite resolution.

\begin{figure*}[t]
\centering
\includegraphics[width=0.95\textwidth]{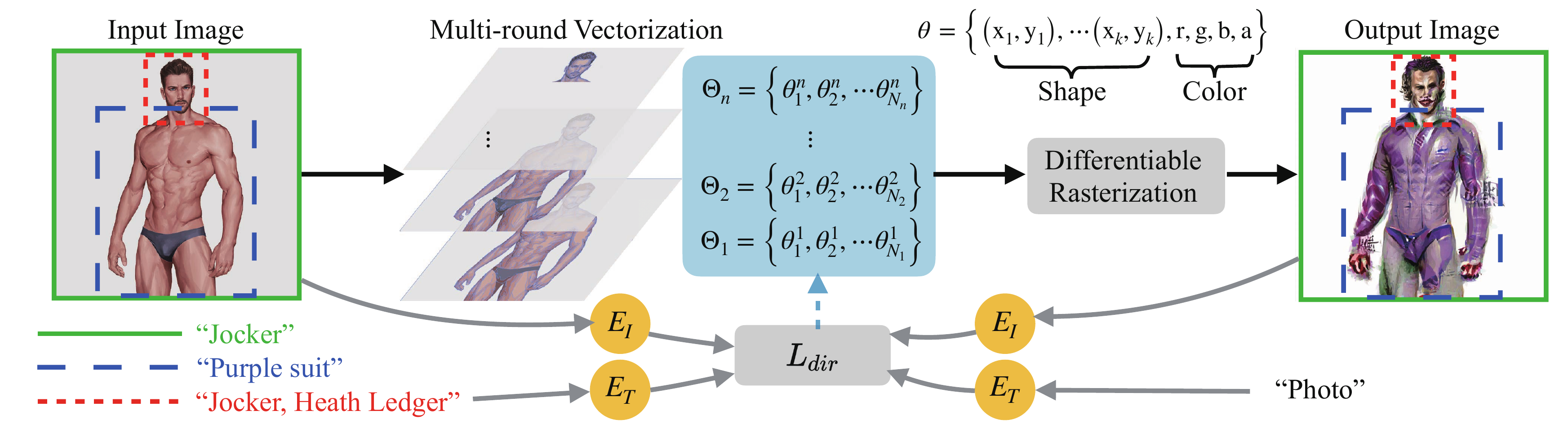} 
\caption{The overall schematics of CLIPVG. We vectorize the input image with a multi-round vectorization strategy. The optimization is guided by an ROI CLIP loss. The parameters are decoupled to enable fine-grained control.}
\label{fig:diagram}
\end{figure*}

\textbf{Text-guided Vector Graphic Generation.}
Diffvg \cite{diffvg} proposes a differentiable rasterizer which supports the raster image based models for the vector graphics, including the CLIP models. The differentiable rasterizer backpropagates the gradients on pixels to the continuous vector graphic parameters, such as the control points and the color. The discrete topology, e.g., the number of vector graphical elements and the connection between the control points, are not changed. We use this topology preserving property of Diffvg to regularize our optimization process.

CLIPdraw\cite{clipdraw} combines CLIP and Diffvg for the first time, and uses CLIP to guide a set of randomly initialized strokes according to the text prompt. StyleCLIPdraw\cite{styleclipdraw} further controls the style of the generated vector graphic by a style image. ES-CLIP \cite{clippaint} uses triangles instead of strokes as the vector graphical elements, and optimizes the triangles using evolution strategy.

The above methods generate the vector graphics from randomly placed vector graphical elements. In contrast, we focus on the manipulation of an existing image.

\textbf{Image Vectorization.}
Raster image vectorization or image tracing is a well-studied problem in computer graphics. Adobe illustrator, the most advanced commercial vector graphic design tool, provides an image tracing tool \cite{AdobeImageTrace} with various control modes and options. By default, Adobe Image Trace (AIT) converts the raster image to a set of non-overlapping filled curves. The vectorization precision can be controlled by the number of target colors. The higher the number of target colors, the higher the precision.

Various other methods are also studied for image vectorization. Direct raster-to-vector conversion with neural networks are supported for the relatively simple images \cite{svg+vae, deepsvg, Im2Vec}.  Stroke-based rendering can be used to fit a complex image with a sequence of vector strokes \cite{learntopaint, stylizedneuralpainting}, but the performance is limited by the predefined strokes. Diffvg can also be leveraged to fit an input image with a set of randomly initialized vector graphical elements. Based on Diffvg, CLIPasso \cite{clipasso} controls the abstraction level or vectorization precision of the output by the number of strokes. LIVE \cite{LIVE} further proposes a coarse-to-fine vectorization strategy.

We adopt a decoupled approach that first vectorizes the input image, then manipulates the vectorized graphic according to the text prompts. Therefore, all of the above methods can be used in our framework. However, the further demand for image manipulation has not been considered by the existing vectorization methods. We introduce a multi-round vectorization strategy that specifically improves the robustness of image manipulation.

\section{Method}
The main schematic of CLIPVG is presented in Figure~\ref{fig:diagram}. We first vectorize the input raster image multiple times with different vectorization precision. All the vector graphical elements are jointly rasterized back to the pixel space by Diffvg. The rasterized image is the reconstruction of the input image at the beginning, and is iteratively optimized towards the direction of the text prompts. The gradients are derived from an ROI (Region Of Interest) CLIP loss and backpropagated to the shape and color parameters of each vector graphical elements. The optimization process is shown in Figure~\ref{fig: iterative_optimization}. More examples can be found in the supplementary material.

\begin{figure}[htb]
\centering
\includegraphics[width=0.47\textwidth]{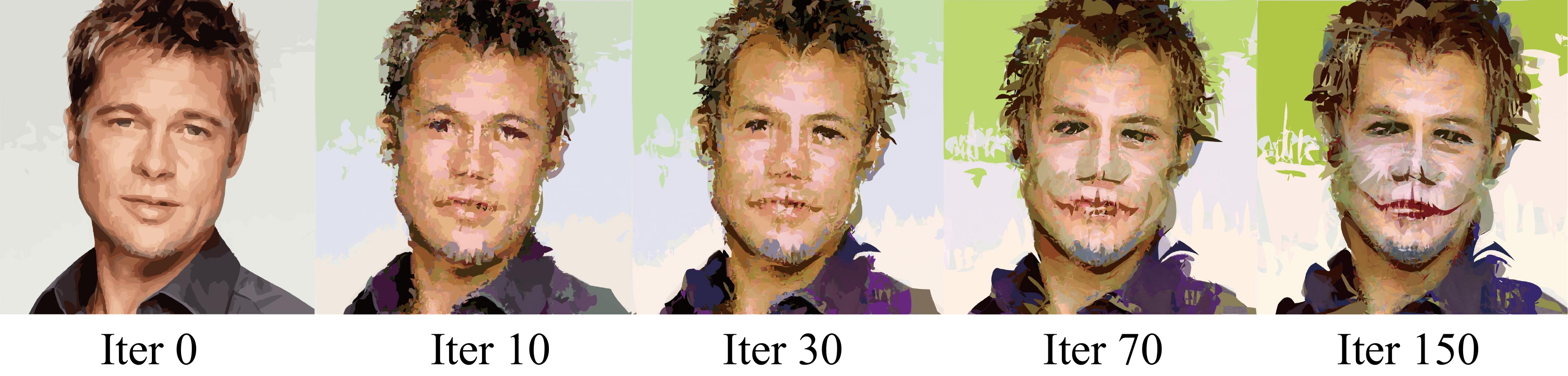} 
\caption{An example of the outputs during the iterative optimization process. the prompt is "Jocker, Heath Ledger".}
\label{fig: iterative_optimization}
\end{figure}

\subsection{Vectorization}
A vector graphic is defined by a set of vector graphical elements. The parameters for each element depends on the type of the element, e.g., a filled curve can be represented as
\begin{equation}
\label{eqn:vector_element}
\theta = \{(x_1, y_1), (x_2, y_2), \dots, (x_m, y_m), r, g, b, a\},
\end{equation}
where $(x_i, y_i)$ are the coordinates of the $i$-th control point. $m$ is the number of control points. $(r, g, b, a)$ are RGB color and opacity values respectively. The optimization on the element is naturally regularized by the constant connection (sequence) of the control points, and the uniform texture within an element.

For an input image, the existing vectorization methods are able to generate a set of vector graphical elements which can be directly optimized by CLIPVG. However, we found several problems with this naive solution. First, although the topology within an element is preserved during optimization, there is no inter-element constraint. Two closely connected elements can be torn apart during the optimization, leaving a gap which is not always desirable. Second, the target image may require extra elements to represent the semantic of the text prompts. For example, the output of "a steampunk style building" should be much more complicated than the input of a cabinet in Figure~\ref{fig:teaser}.
But the generation of new vector graphical elements is non-differentiable and is not supported by the optimization process.

To mitigate the above issues, we propose a multi-round vectorization strategy which takes into consideration the further need of image manipulation. We vectorize the input raster image multiple times with different vectorization precision, and derive a unique set of vector graphical elements from each round of vectorization,
\begin{equation}
\Theta_i = \{\theta_1^i, \theta_2^i, \dots, \theta_{N_i}^i\},
\end{equation}
where $N_i$ is the number of elements, and $\theta_j^i$ is the $j$-th element for the $i$-th round of vectorization. We increase the vectorization precision for each round, which usually results in more elements, i.e., $N_{i+1} > N_i$. We can further enhance a key region, e.g., the human face area in Figure~\ref{fig:diagram}, by another round of vectorization for the specific region.

We combine all the elements by placing the $(i+1)$-th set of elements on top of the $i$-th set of elements. The full parameter set from $n$ rounds of vectorization is
\begin{equation}
\Theta = \{\Theta_1, \Theta_2, \dots, \Theta_n\}.
\end{equation}
Our multi-round strategy can be used to enhance any existing vectorization method. The additional vector graphical elements allow CLIPVG to generate finer details according to the prompts. Moreover, the gap between the vector graphical elements can be filled by the redundant elements.

\subsection{Loss Function}

Similar to \cite{stylegannada, diffusionclip, clipstyler}, we adopt a directional CLIP loss which is defined to align the latent directions of the text and images,
\begin{equation}
\label{eqn:ldir}
\begin{array}{r}
\Delta T=E_{T}\left(t_{pr}\right)-E_{T}\left(t_{ref}\right), \vspace{1ex}\\
\Delta I=E_{I}\left(I_{gen}\right)-E_{I}\left(I_{src}\right), \vspace{1ex}\\
L_{dir}(t_{pr},t_{ref}, I_{gen}, I_{src})=1-\frac{\Delta I \cdot \Delta T}{|\Delta I || \Delta T|},
\end{array}
\end{equation}
where $t_{pr}$ is the text prompt. $t_{ref}$ is a reference text which is fixed to "photo" in our implementation. 
$I_{gen}$ is the generated image which is to be optimized. $I_{src}$ is the source images. $E_T$ and $E_I$ are the text and image encoders of CLIP. $\Delta T$ and $\Delta I$ are the latent directions of the text and the images respectively. We will neglect the fixed $t_{ref}$ and denote the loss as $L_{dir}(t_{pr}, I_{gen}, I_{src})$ in the following analysis.

We support multiple input text prompts, each associated with a specific ROI as shown in Figure~\ref{fig:diagram}. The ROI CLIP loss is
\begin{equation}
\label{eqn:roi_loss}
L_{dir}(\Theta, t_{pr}^i, A_i) = L_{dir}(t_{pr}^i, C_{A_i}(R(\Theta)), C_{A_i}(I_{init})),
\end{equation}
where $A_i$ is the area of the $i$-th ROI, and $t_{pr}^i$ is the associated prompt. $R$ is the differentiable rasterizer. $R(\Theta)$ is the rasterized image. $I_{init}$ is the input raster image. $C_{A_i}(I)$ is an operation to crop the area $A_i$ from the image $I$.

Our ROI CLIP loss in Eq. \ref{eqn:roi_loss} is also generalized to support the random cropping enhancement in CLIPstyler \cite{clipstyler}. In this case, The CLIP loss is applied to a number of patches which are randomly cropped from the output image. This approach was adopted by CLIPstyler to enhance the local texture, and is further leveraged by our framework as an data augmentation method. For each ROI which is directly associated with an input text prompt, we apply the random cropping enhancement to derive a number of patches which are associated with the same prompt. We then calculate the ROI CLIP loss for each patch according to Eq.~\ref{eqn:roi_loss}. The total loss of CLIPVG is
\begin{equation}
\label{eqn:ltotal}
L_{total}=\sum_{i=1}^{h} w_{i}L_{dir}(\Theta, t_{pr}^i, A_i),
\end{equation}
where $w_i$ is the weight of CLIP loss for the $i$-th region. $h$ is the total number of regions. A region can be either an ROI, or a patch which is randomly cropped from an ROI.

\subsection{Optimization}
The set of parameters $\Theta$ is optimized to minimize the total loss in Eq.~\ref{eqn:ltotal} using Diffvg \cite{diffvg}. The shape and color parameters are naturally decoupled in Eq.~\ref{eqn:vector_element}, where the shape is defined by the control points, and the color is defined by the RGB and opacity values. Therefore, we can optimize the shape and color parameters independently with two different learning rates. This is especially useful to keep either the shape or the color unchanged.

We can also edit only a subset of the vector graphical elements. The subset is usually defined by the elements which initially intersect with a certain subregion. Our framework generally allows the editable elements to move partially or fully outside the subregion during the iterative optimization process, leading to a seamless connection between the subregion and the rest of the image.

\section{Experiments}
\subsection{Experiment Setup}

\textbf{Implementation.} The multi-round vectorization strategy of CLIPVG requires an arbitrary vectorization tool, e.g., AIT \cite{AdobeImageTrace}, Diffvg \cite{diffvg}, LIVE \cite{LIVE}, etc. We use AIT \cite{AdobeImageTrace} as the default tool, since it gives the most accurate reconstruction results in our experiments. We adopt two rounds of vectorization by default. The first round is done by $N_c=10$, and the second round is done by $N_c=30$, where $N_c$ is the number of target colors in AIT. We add another round of vectorization for the area of human face with $N_c=30$.

We apply random cropping to obtain $N_{patch}=64$ patches from each ROI. The patches are randomly cropped in each iteration. The default CLIP loss weight is 30.0 for a text prompt associated ROI, and is $80.0/N_{patch}$ for each randomly cropped patch. The patch size is always set to $80\%$ the longer edge of the ROI region, e.g., $400\times400$ for a $500\times300$ ROI, and zero-padding is adopted when necessary. Similar to CLIPstyler \cite{clipstyler}, we also apply the random perspective augmentation to the patches.

Similar to \cite{clipstyler, styleclip}, we use the ViT-B/32 CLIP model \cite{CLIP}. We employ the Adam \cite{adam} optimizer with a learning rate of 0.2 for the shape parameters, and 0.01 for the color parameters by default. The number of iterations is set to 150. The running time information is included in the supplementary material.

\textbf{Baseline Methods.} There is no existing text-guided manipulation method for the vector graphics. So we mainly compare to the state-of-the-art raster image based methods. We consider two domain-agnostic baselines, Disco Diffusion v5.6 \cite{DiscoDiffusion} and CLIPstyler \cite{clipstyler}. Disco Diffusion is a popular open-source project based on a general diffusion model. CLIPstyler is a CLIP-guided style transfer method which does not rely on any generative model. We also compare CLIPVG to three domain specific methods, including StyleCLIP \cite{styleclip}, StyleGAN-NADA \cite{stylegannada} and DiffusionCLIP \cite{diffusionclip}. The first two are based on the StyleGAN models, while the last one is diffusion model based. We run all the baseline methods with the official code base and the default configuration. We use images with a resolution of $512\times512$ as the inputs.

\begin{figure}[t]
     \centering
     \begin{subfigure}{\linewidth}
         \centering
         \includegraphics[width=\linewidth]{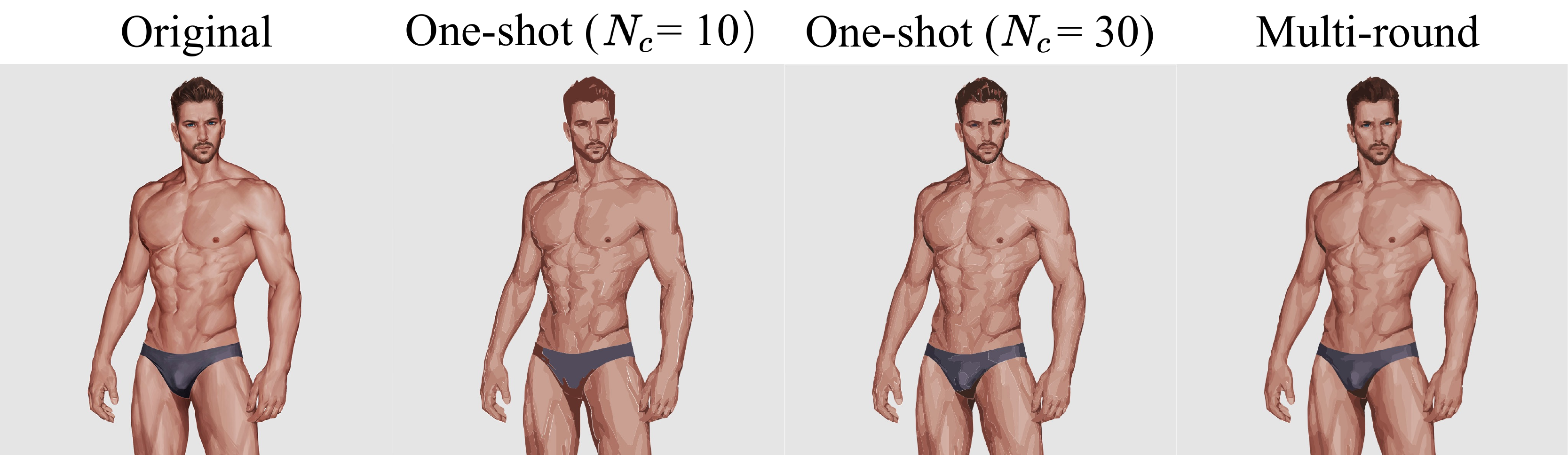}
         \caption{The initialized vector graphics}
         \label{fig:mutilayer_vectorized}
    \end{subfigure}
    \hfill
     \begin{subfigure}{\linewidth}
         \centering
         \includegraphics[width=0.8\linewidth]{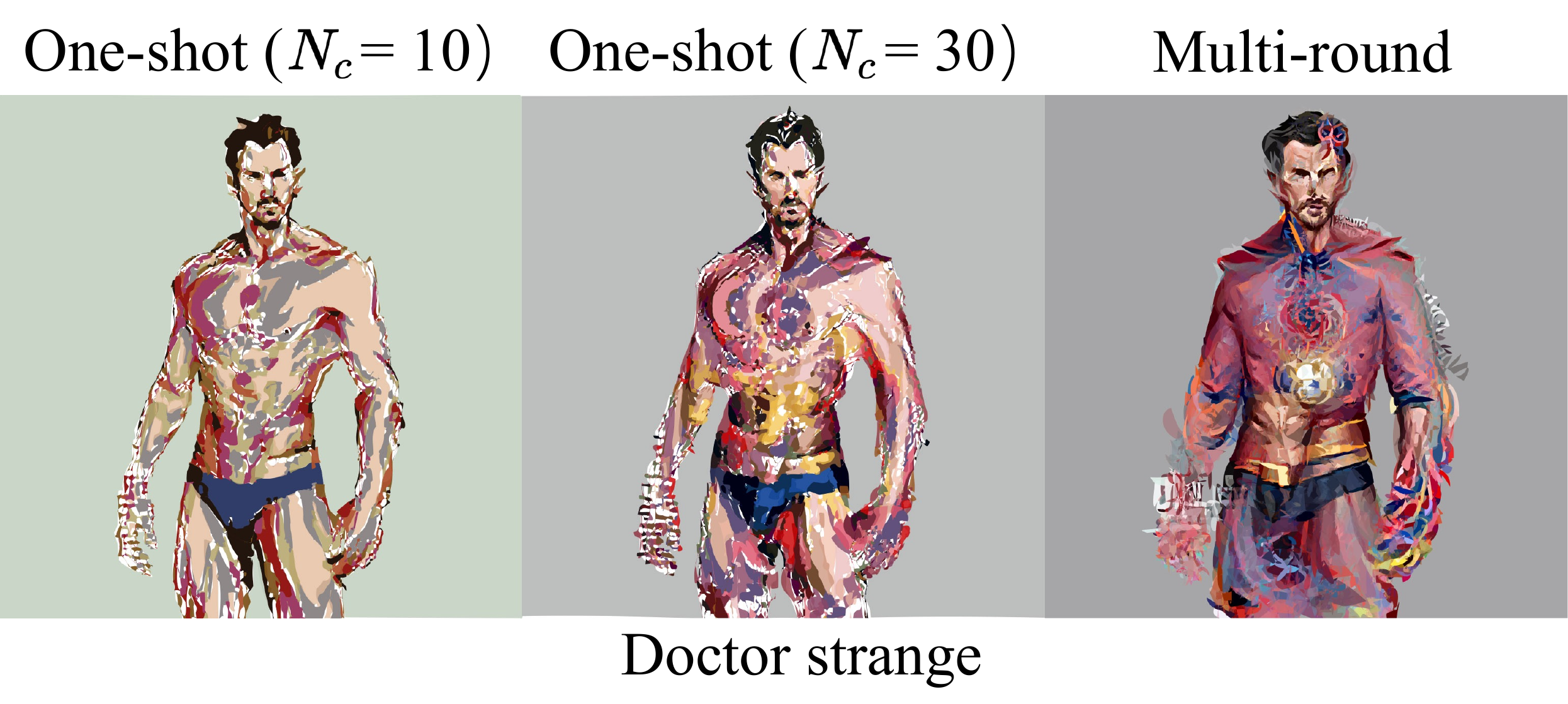}
         \caption{The manipulation results with different vectorization strategies.}
         \label{fig:mutilayer_manipulated}
    \end{subfigure}
\caption{Image vectorization and manipulation results with and without the multi-round strategy.}
\label{fig:multilayer}
\end{figure}

\begin{figure}[t]
     \centering
     \begin{subfigure}{\linewidth}
         \centering
         \includegraphics[width=\linewidth]{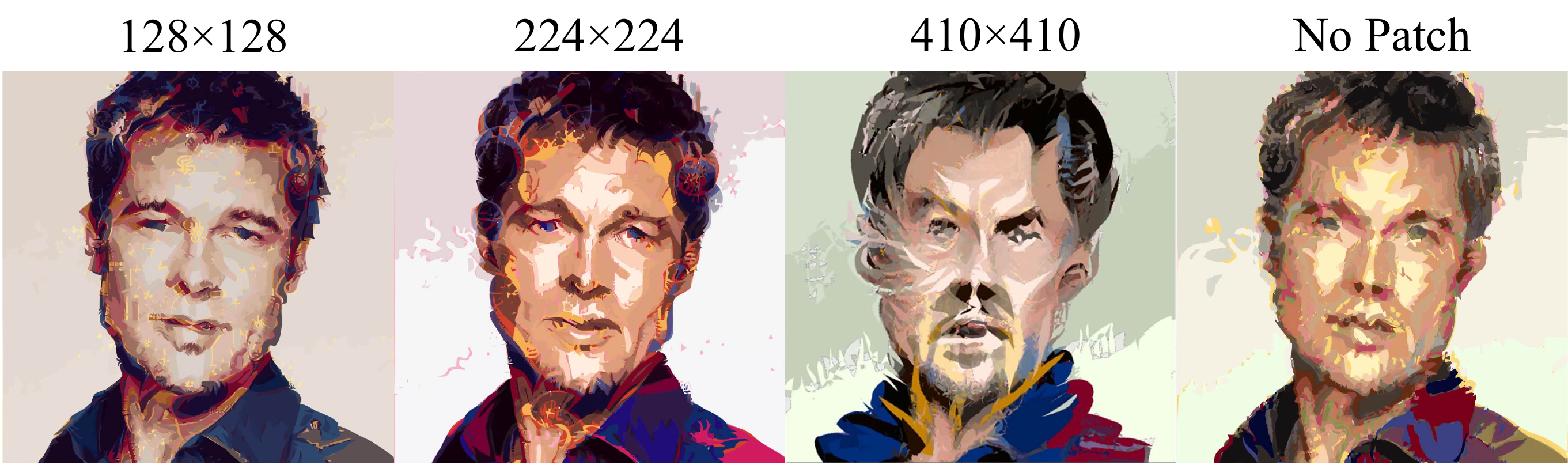}
         \caption{CLIPVG results with different patch configurations.}
         \label{fig:patch_clipvg}
    \end{subfigure}
    \hfill
     \begin{subfigure}{\linewidth}
         \centering
         \includegraphics[width=\linewidth]{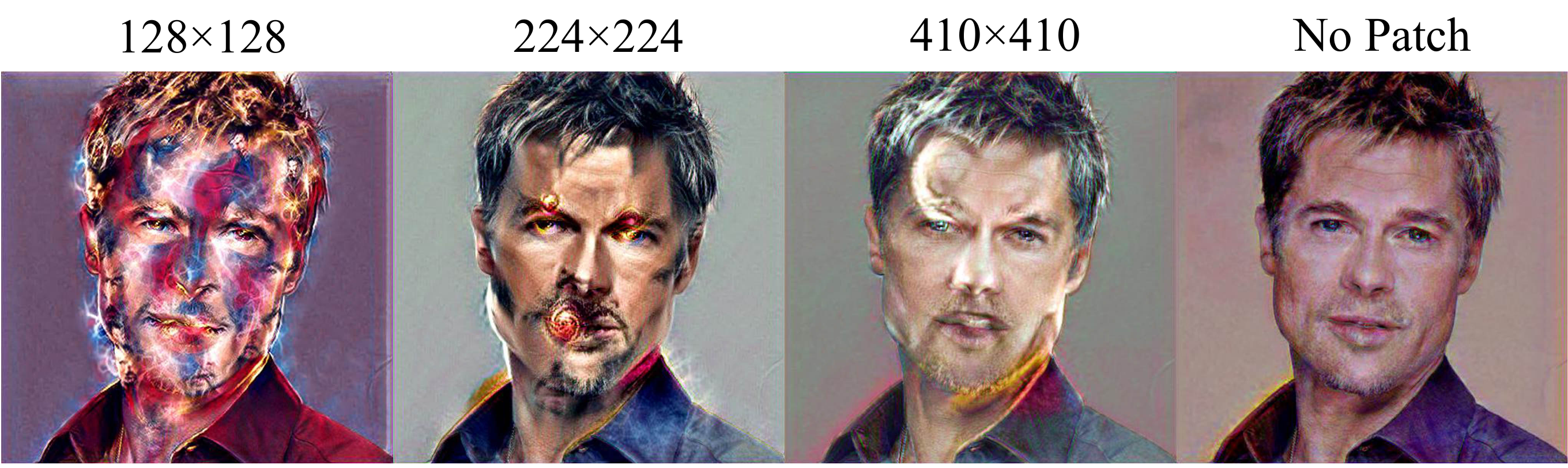}
         \caption{CLIPstyler results with different patch configurations.}
         \label{fig:patch_clipstyler}
    \end{subfigure}
\caption{CLIPVG and CLIPstyler results with different patch size configurations for the random cropping enhancement. The prompt is "Doctor Strange".} 
\label{fig:patch}
\end{figure}

\subsection{Ablation}
\textbf{Multi-Round Vectorization.} We compare our multi-round vectorization strategy to the one-shot methods in Figure~\ref{fig:multilayer}. The one-shot vectorization is done by AIT with $N_c=10$ and $N_c=30$ respectively. The multi-round vectorized image consists of all the elements from the one-shot cases, plus an additional set of elements for the face part of the image. It can be seen from Figure~\ref{fig:mutilayer_vectorized} that the multi-round vectorization strategy achieves the best reconstruction precision due to the availability of more vector graphical elements. After the image manipulation, some undesirable white spaces appear in the one-shot cases, since these areas are not covered by any elements. The white space issue is greatly alleviated by the multi-round strategy as shown in Figure~\ref{fig:mutilayer_manipulated}. Furthermore, the manipulated image of the multi-round case also has richer details than the one-shot cases.

\begin{figure}[t]
\centering
\includegraphics[width=1.0\linewidth]{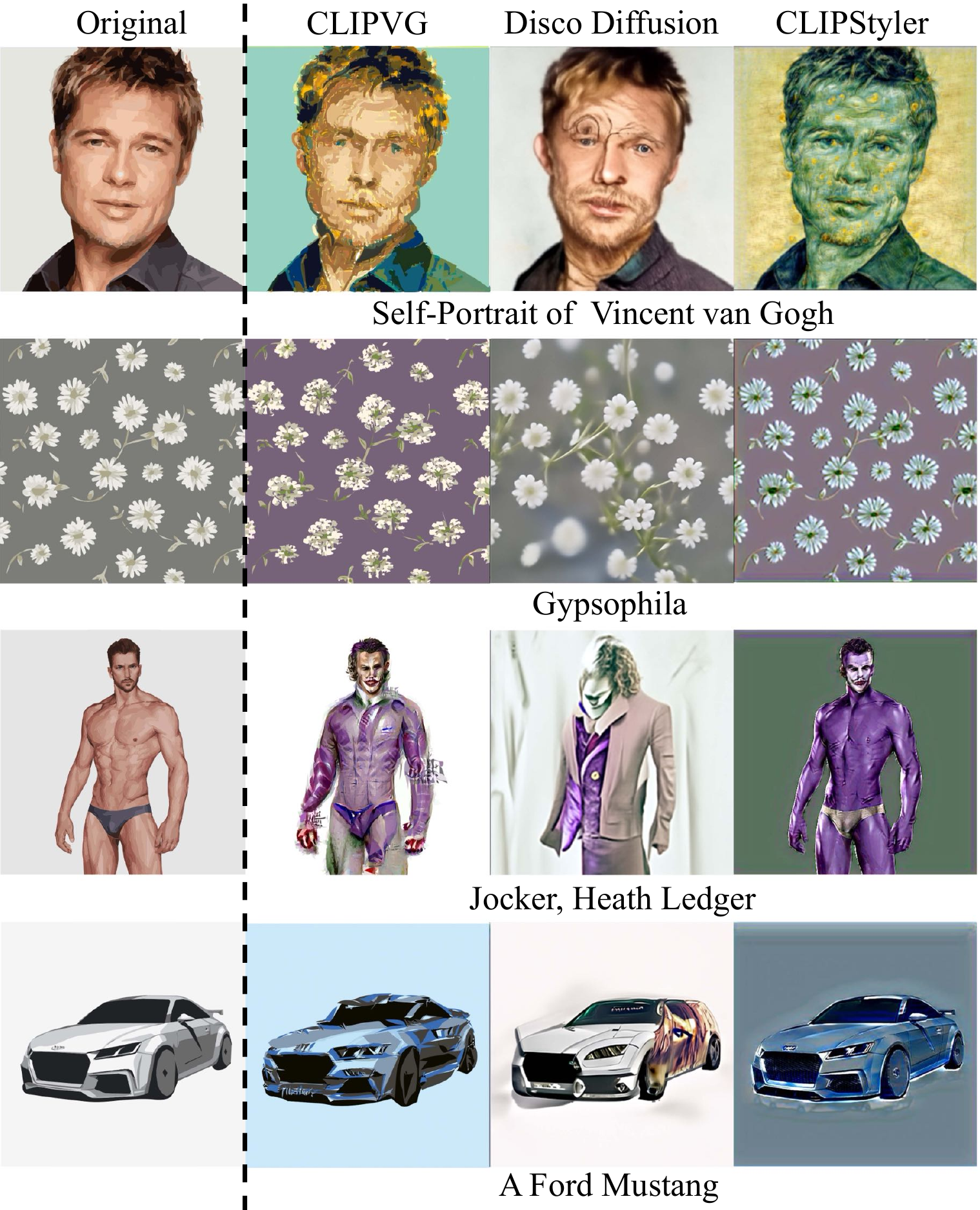} 
\caption{Comparison with the domain-agnostic methods.}
\label{fig:cmp_domain_agnostic}
\end{figure}

\textbf{Random Cropping Enhancement.} We also evaluate the effect of different random cropping configurations in Figure~\ref{fig:patch}. We try the patch size of $128\times128$, $224\times224$, $410\times410$, or no random cropping. Note that CILPstyler~\cite{clipstyler} has done a similar study and chosen $128\times128$ as the default patch size. We revisit the experiment here for two purposes. First, the text prompts used by CLIPstyler are generally related to the low level texture, e.g., "a cubism style painting". But we consider the prompts which require high-level semantic manipulation, e.g., "Doctor Strange" in Figure~\ref{fig:patch}. Second, the low level texture can be constrained by the vector graphic specific regularization instead of the patch-wise loss in our case.

We show the results of CLIPVG in Figure~\ref{fig:patch_clipvg}, and the results of CLIPstyler in Figure~\ref{fig:patch_clipstyler} as a reference. The original image is the same as the first row in Figure~\ref{fig:teaser}. When the patch size is relatively small, i.e., $128\times128$ or $224\times224$, We can see some obvious local artifacts with Doctor Strange's classic red-and-blue color scheme for both CLIPVG and CLIPstyler. The artifacts indicate that a small patch size is not feasible for the high-level semantic manipulation tasks.

The local artifacts are suppressed by using a large patch size or disabling the random cropping. CLIPVG with a patch size of $410\times410$ can effectively change the hairstyle and the identity of the face according to the prompt. CLIPVG without random cropping suffers from less accurate semantic change and blurry details, indicating that the random cropping enhancement is still beneficial for the overall quality.
On the other hand, there is little meaningful semantic change for CLIPstyler even if the patch is enlarged or disabled. It only achieves limited skin color or local texture change, implying that the optimization is stuck in a local optimum due to the lack of low level constraints.

In conclusion, a large patch size can be adopted by our vector graphic based optimization process to achieve robust high-level semantic manipulation. As a result, we relax the constraints on the small patches, and select a larger default patch size than CLIPstyler, i.e., $410\times410$ for the region of $512\times512$.

\begin{figure}[t]
\centering
\includegraphics[width=1.0\linewidth]{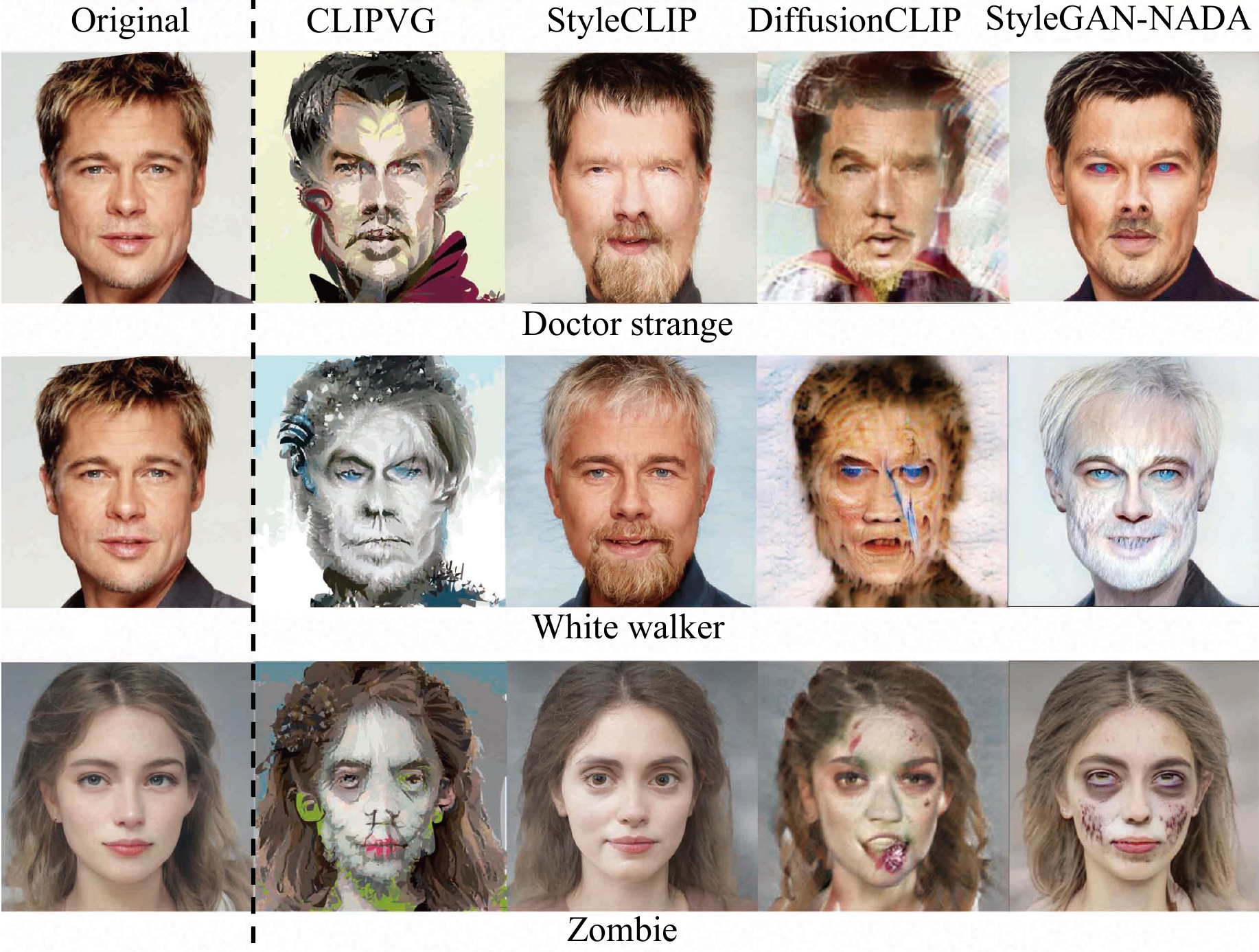} 
\caption{Comparison with the domain-specific Methods.}
\label{fig:cmp_domain_specifc}
\end{figure}

\begin{figure}[b]
     \centering
     \begin{subfigure}[b]{0.23\textwidth}
         \centering
         \includegraphics[width=\textwidth]{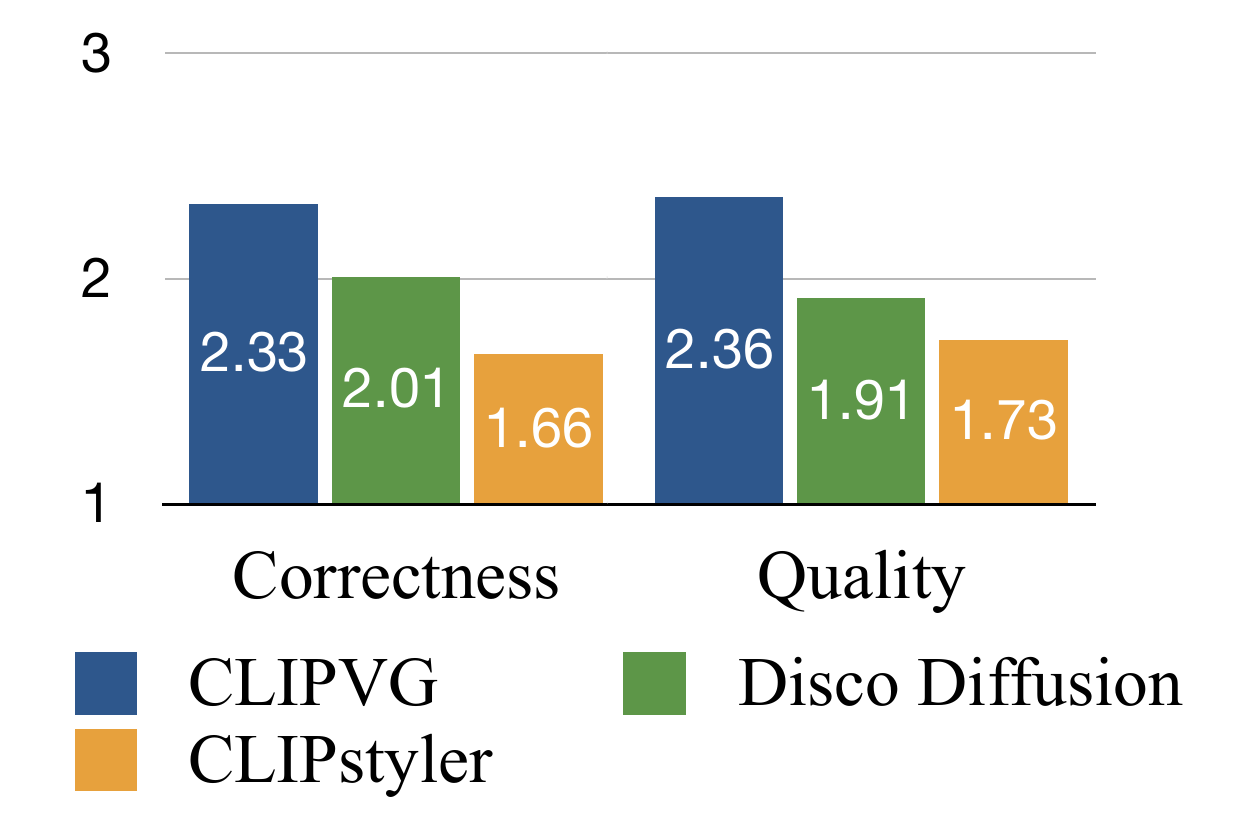}
         \caption{Domain-agnostic.}
         \label{fig:pilot_agnostic}
    \end{subfigure}
    \hfill
     \begin{subfigure}[b]{0.23\textwidth}
         \centering
         \includegraphics[width=\textwidth]{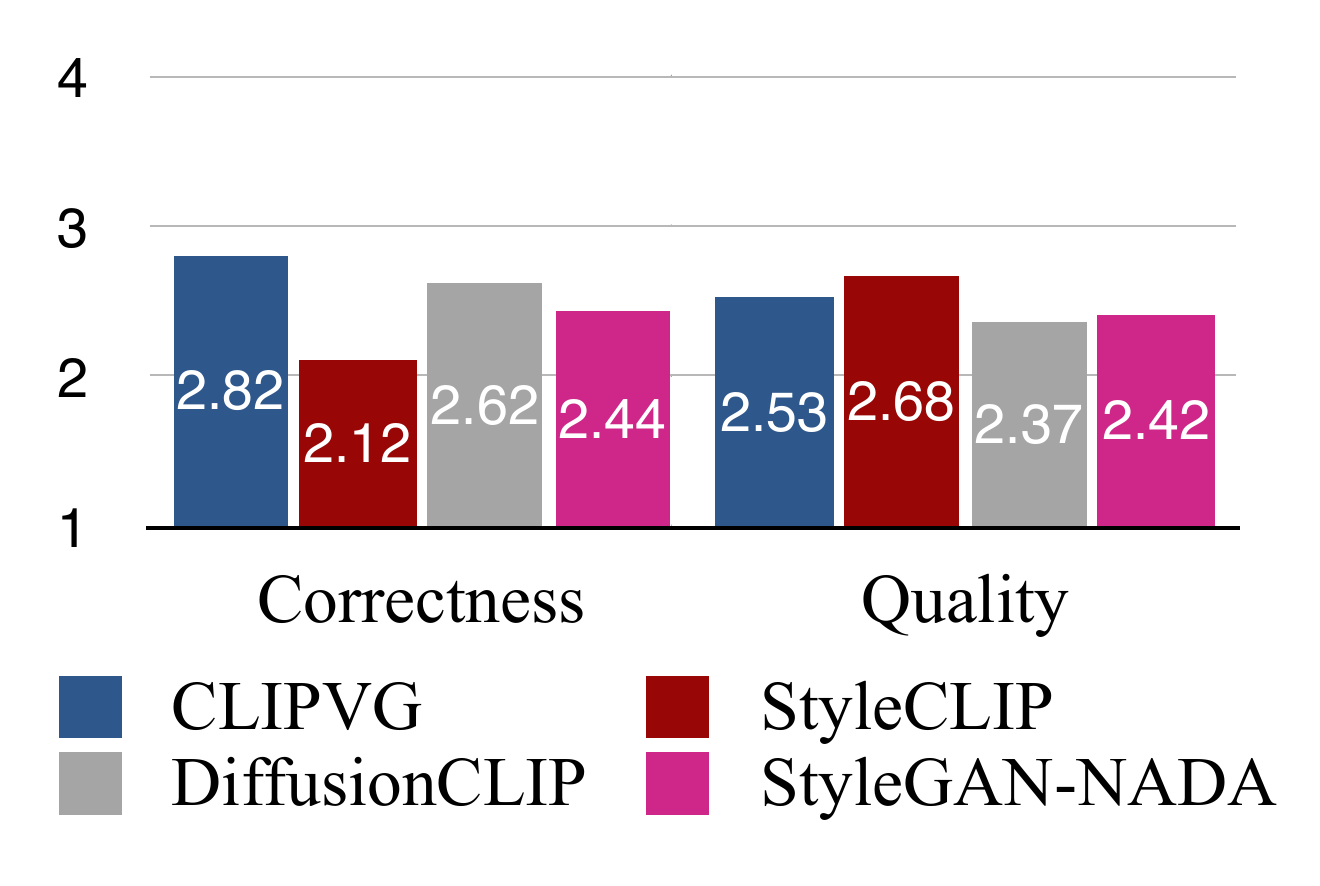}
         \caption{Domain-specific.}
         \label{fig:pilot_specific}
    \end{subfigure}
\caption{Average user ratings of different methods in the pilot study.} 
\label{fig:pilot}
\end{figure}

\begin{figure*}[htb]
     \centering
     \begin{subfigure}{0.33\textwidth}
         \centering
         \includegraphics[width=\textwidth]{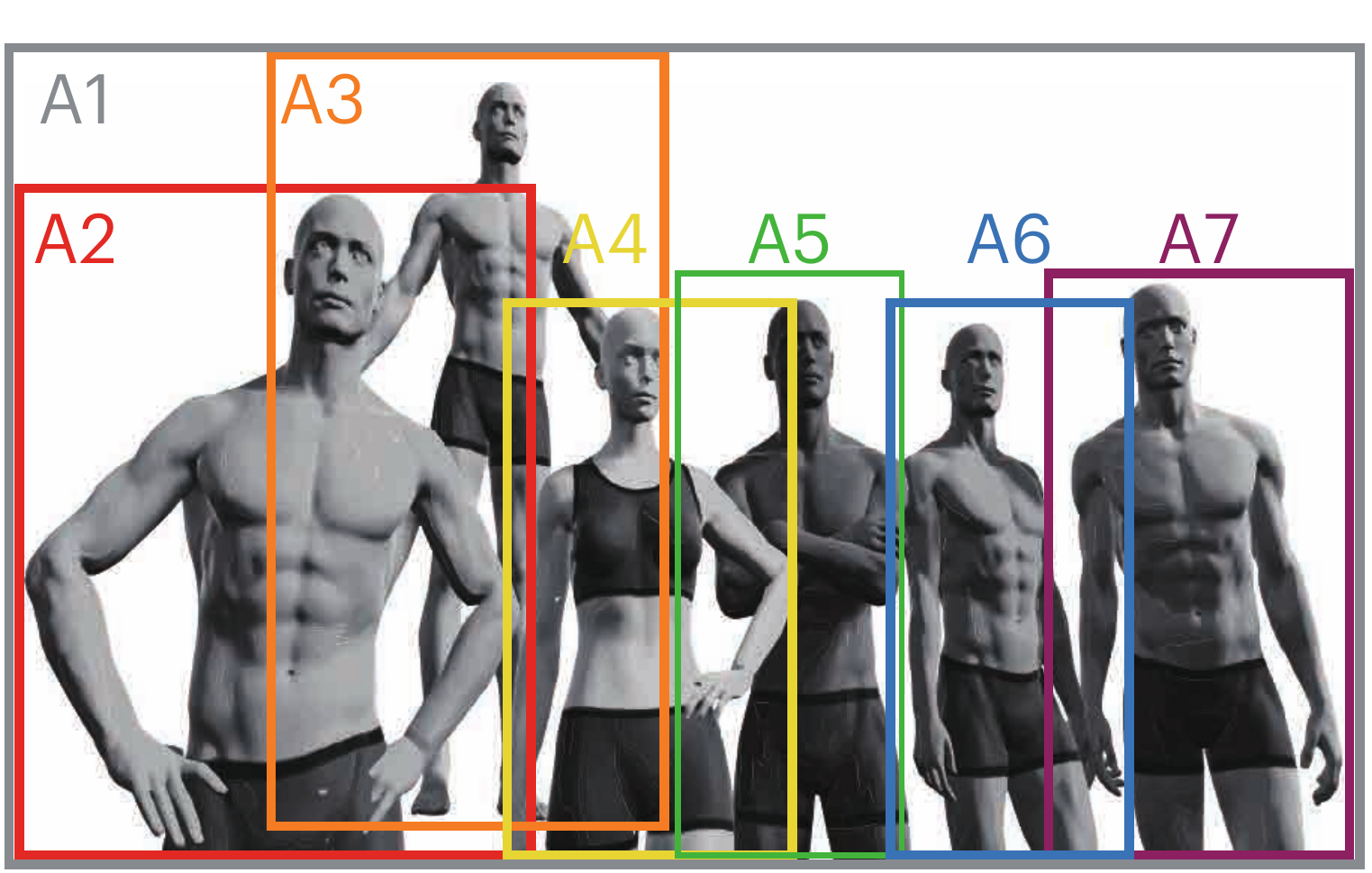}
         \caption{The original image and the ROIs.}
         \label{fig:dc_original}
    \end{subfigure}
    \hfill
     \begin{subfigure}{0.33\textwidth}
         \centering
         \includegraphics[width=\textwidth]{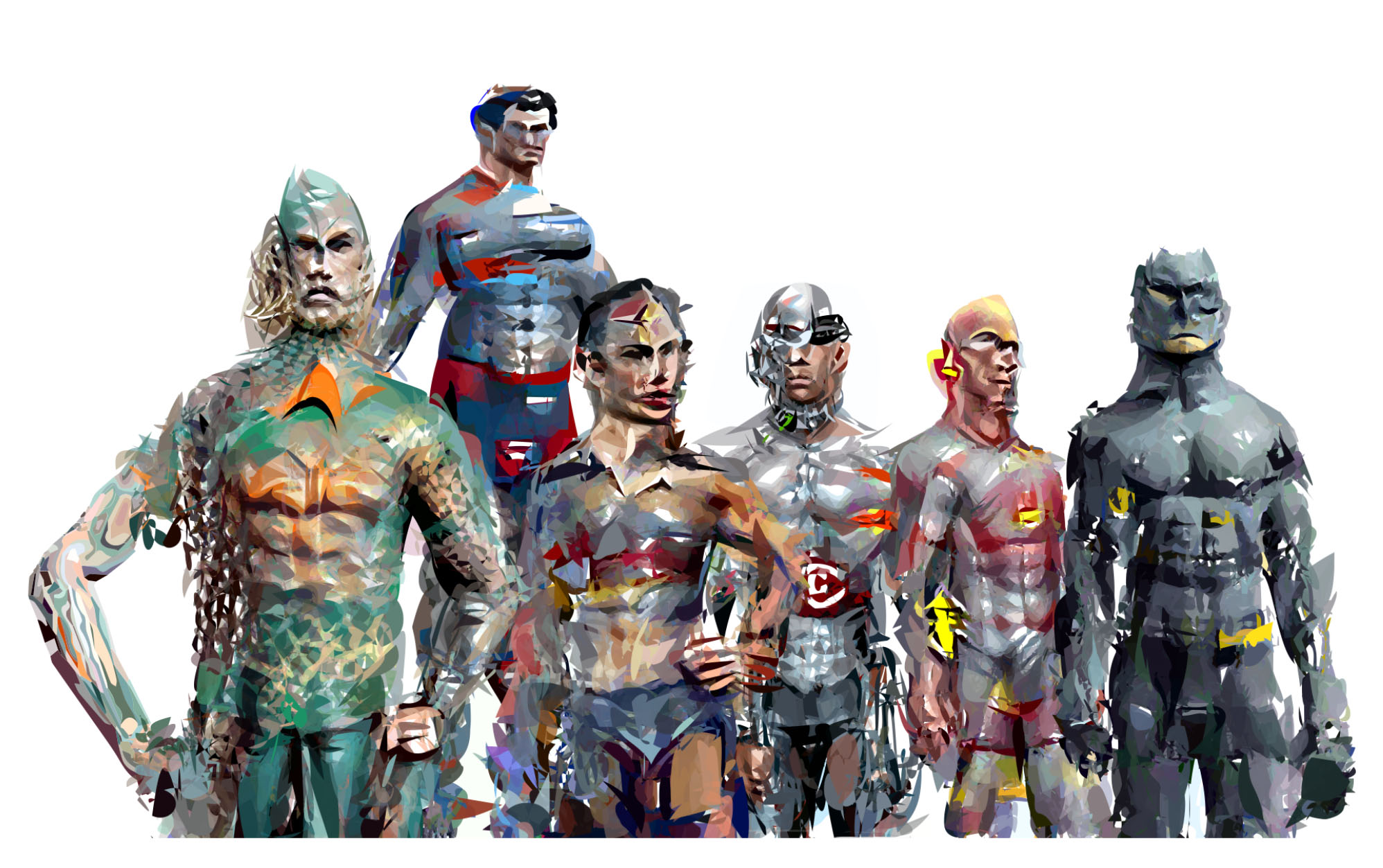}
         \caption{Output with all the ROI prompts.}
         \label{fig:dc_roi}
    \end{subfigure}
    \hfill
     \begin{subfigure}{0.33\textwidth}
         \centering
         \includegraphics[width=\textwidth]{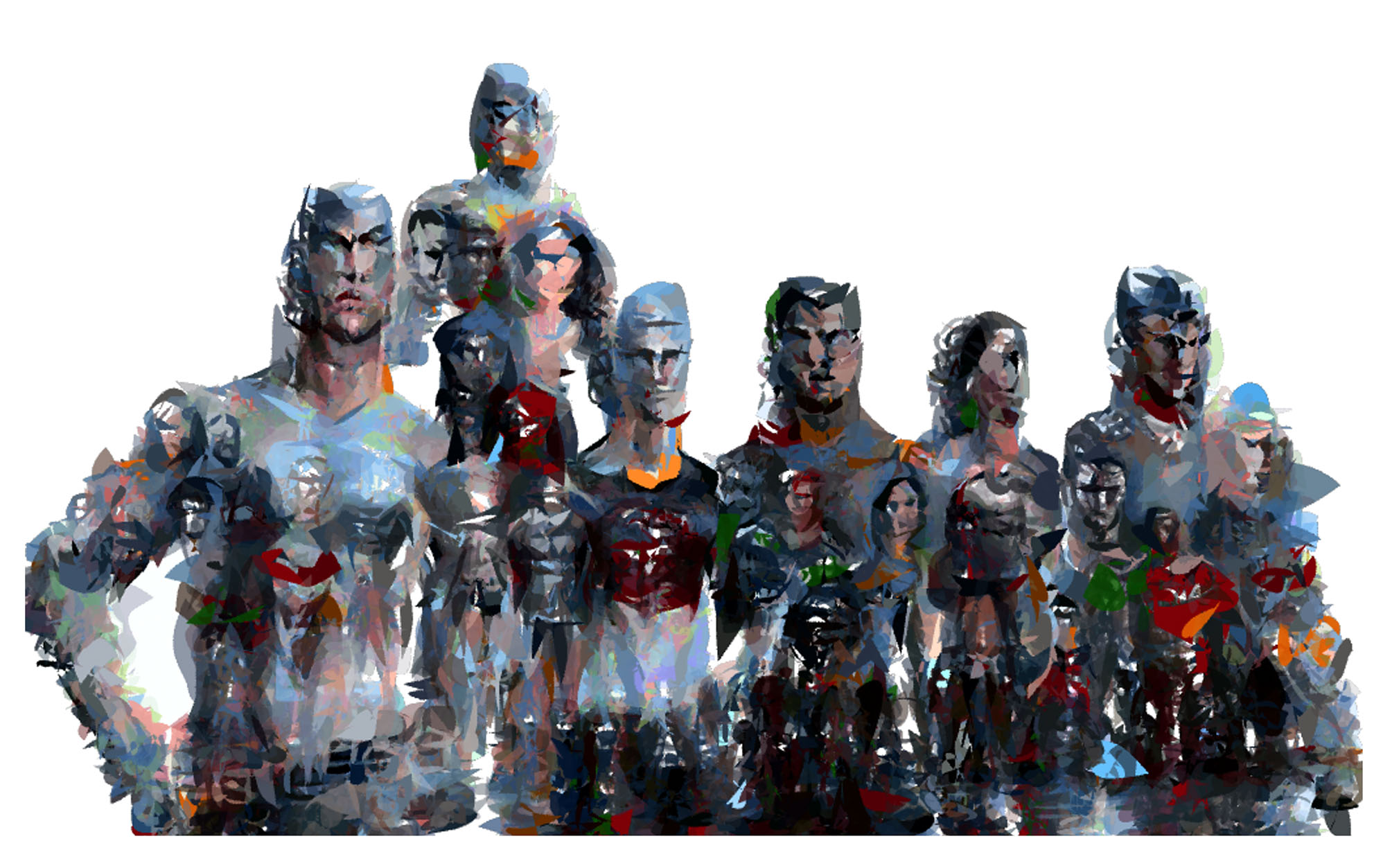}
         \caption{Output with the global prompt only.}
         \label{fig:dc_global}
    \end{subfigure}
\caption{Text-guided manipulation results with and without the ROI prompts. The prompts are: 1. "Justice League Six", 2. "Aquaman", 3. "Superman", 4. "Wonder Woman", 5. "Cyborg", 6. "Flash, DC Superhero" and 7. "Batman". The areas of each ROI are shown as A1 to A7 in (a). The global prompt of (c) is the concatenation of all the prompts.} 
\label{fig:multi_prompts}
\end{figure*}

\begin{figure}[h!]
     \centering
     \begin{subfigure}{\linewidth}
         \centering
         \includegraphics[width=\textwidth]{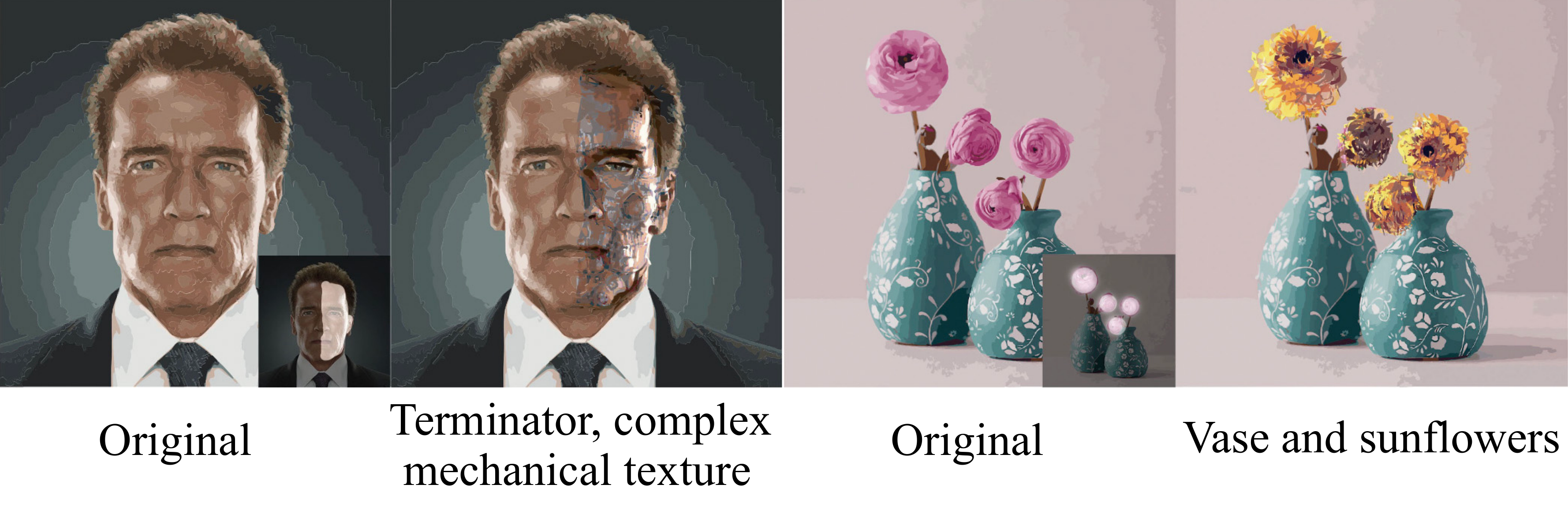}
         \caption{Subregion editing.}
         \label{fig:roi_edit}
    \end{subfigure}
    \hfill
     \begin{subfigure}{\linewidth}
         \centering
         \includegraphics[width=\textwidth]{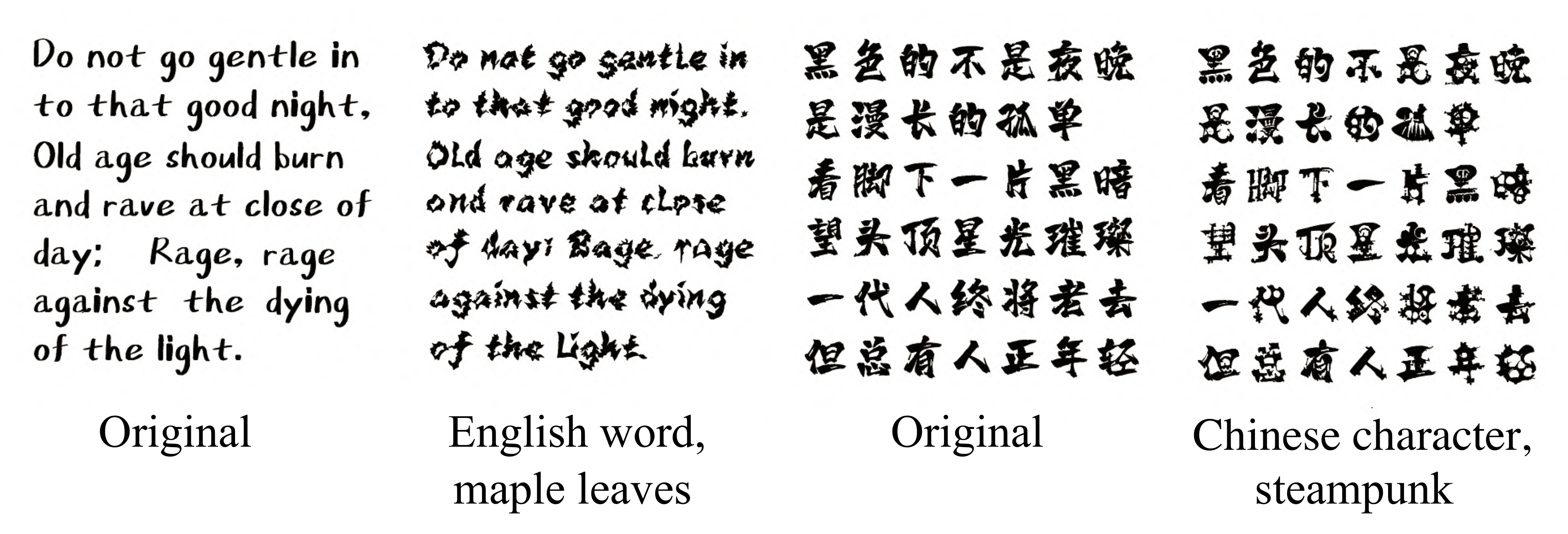}
         \caption{Font stylization.}
         \label{fig:shape_edit}
    \end{subfigure}
    \hfill
     \begin{subfigure}{\linewidth}
         \centering
         \includegraphics[width=\textwidth]{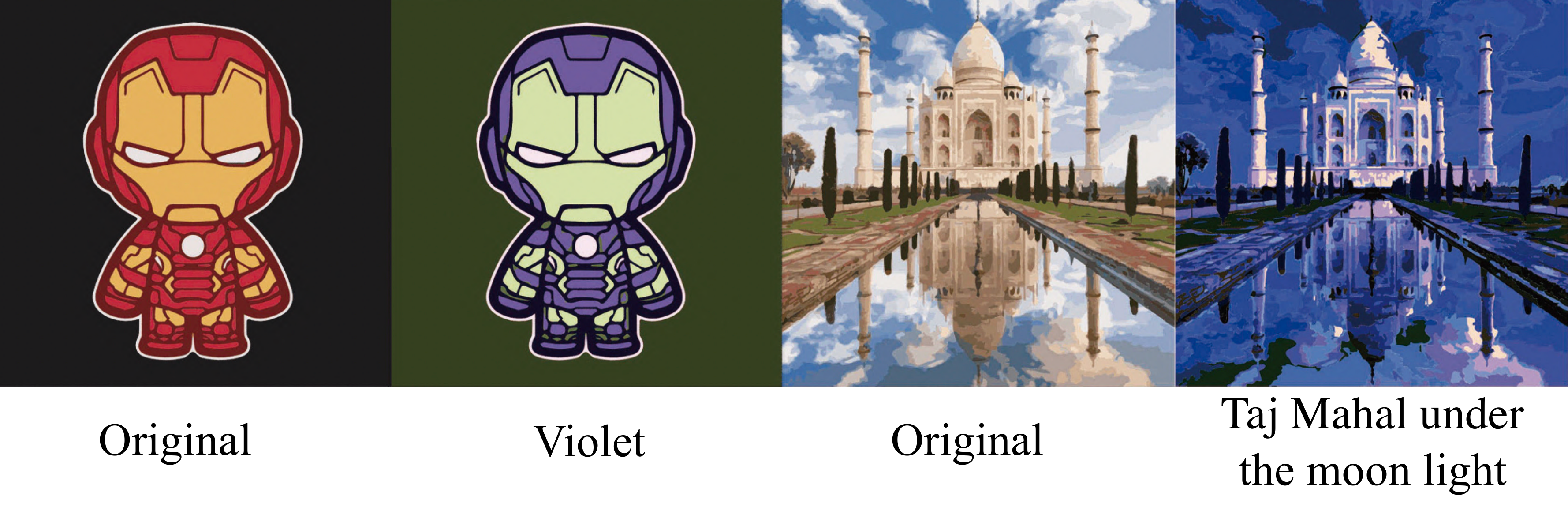}
         \caption{Re-colorization.}
         \label{fig:color_edit}
    \end{subfigure}
\caption{Separate control of parameters. (a) optimizes the parameters of a specific subregion. The target subregion is shown in the bottom right corner of the input. (b) optimizes the shape parameters. (c) optimizes the color parameters.} 
\label{fig:param_decouple}
\end{figure}

\subsection{Comparisons}
\textbf{Domain-Agnostic Methods.} We compare CLIPVG to Disco Diffusion and CLIPstyler in Figure~\ref{fig:cmp_domain_agnostic}.  CLIPVG generally delivers the desired semantic transfer. It is the only method which achieves both identity and texture change in the "Self-Portrait of Vincent van Gogh" case, while the other methods only address one aspect. It also manages to modify the number and shape of petals according to "Gypsophila". The results of Disco Diffusion are relatively unstable, e.g., the position of head is incorrect in the "Joker, Heath Ledger" case. It also generates a head of the Mustang Horse in the "A Ford Mustang" case, which can be taken as a local optimal solution from the pixel-level manipulation. The results indicate that domain agnostic image manipulation is a very challenging problem even with the help of a large-scale pre-trained generative model. CLIPstyler is always limited to the local texture and color change due to the strict patch-wise constraints. Compared to the raster image based methods, CLIPVG can focus on the global semantic and suppress the over-editing of the local area by leveraging the vector graphic specific regularization.

\textbf{Domain-Specific Methods.} We compare CLIPVG to StyleCLIP, StyleGAN-NADA and DiffusionCLIP in Figure~\ref{fig:cmp_domain_specifc}. Similar to the domain-agnostic case, the results of CLIPVG correctly reflect the semantics of the text prompts. StyleCLIP generally fails to match the text prompt when the desired change is out of the domain of the original StyleGAN model. StyleGAN-NADA and DiffusionCLIP can better change the semantics by finetuning the generative model according to the prompts. The semantic manipulation of CLIPVG is sometimes more thorough than StyleGAN-NADA and DiffusionCLIP, e.g., the color of cloth is changed more accurately according to the "Doctor Strange" prompt. This can be explained as StyleGAN-NADA and DiffusionCLIP tend to produce results based on some general domain knowledge learned from a set of images, not respecting each individual input image as mush as CLIPVG does.

\textbf{Quantitative Results.} We conduct a pilot study for quantitative comparison. The users are asked to compare different methods from two perspectives, the semantic correctness, i.e., how well do the output image and the prompt fit together, and the image quality. The details can be found in the supplementary document.

The domain-agnostic and domain-specific results are shown in Figure~\ref{fig:pilot_agnostic} and \ref{fig:pilot_specific} respectively. CLIPVG outperforms Disco Diffusion and CLIPstyler for both the semantic correctness and the image quality in Figure~\ref{fig:pilot_agnostic}, and achieves comparable performance as the state-of-the-art domain-specific methods in Figure~\ref{fig:pilot_specific}. The pilot study confirms the strong semantic manipulation capability and the robustness of CLIPVG, which come from the vector graphic specific regularization and the multi-round vectorization.

Notice that, compared with other domain-specific methods, StyleCLIP tends to produce results of higher quality but of weaker semantic connections with the given text prompts. This is because StyleCLIP strictly constrains the output to be within the  domain of the pre-trained StyleGAN model. The performances of StyleGAN-NADA and DiffusionCLIP are more balanced because they finetune the models based on input text prompts, while CLIPVG naturally achieves such balance without additional efforts.

In addition to the pilot study, we also present the quantitative results of CLIP score \cite{glide, diffusionclip, clipstyler} in the supplementary material for the further evaluation of the semantic correctness.

\subsection{Fine-Grained Control}

Our flexible framework supports a set of fine-grained control methods, including the ROI specific prompts and the separate control of parameters. These methods can be applied to various applications on a needed basis.

\textbf{ROI Prompts.} CLIPVG is domain-agnostic and is not restricted by any generative model. It can be potentially leveraged to manipulate a complicated image containing multiple objects. The ROI CLIP guidance allows us to define different targets for the different objects. An example is shown in Figure~\ref{fig:multi_prompts}, where each character is assigned a different prompt. It can be seen from Figure~\ref{fig:dc_roi} that when ROI prompts are enabled, each character in the output image has a clear correspondence with its associated prompt. In contrast, the identity of each character becomes very ambiguous if a global prompt is used instead of the ROI prompts, as shown in Figure~\ref{fig:dc_global}. It is worth noting that transforming each character separately by a domain-specific method is not practical, since the ROIs are overlapping in this example.

\textbf{Separate Control.} 
CLIPVG optimizes the shape and color of all the vector graphical elements simultaneously by default. But it is sometimes desirable to edit a certain subregion or a certain aspect of the image. We can define a target subregion by a mask, and optimize the vector graphical elements within the subregion as shown in Figure~\ref{fig:roi_edit}. We stylize the fonts in Figure~\ref{fig:shape_edit} by editing the shape of the elements and keeping the color unchanged. Re-colorization is done in Figure~\ref{fig:color_edit} by optimizing only the color parameters.

\section{Conclusion}
We introduce CLIPVG, the first vector graphic based solution for text-guided image manipulation. The optimization process is greatly stabilized by the vector graphic specific regularization. We eliminate the dependency on additional pre-trained models, and support domain-agnostic image manipulation. We develop a robust multi-round vectorization strategy, and a set of fine-grained control methods which enables a wide range of applications. Extensive experiments and human evaluation confirm the superior semantic transfer performance and robustness of our method over the existing baselines.



\newpage

\section{Supplementary Material}

In this document, we first describe the details of the pilot study, followed by additional evaluation results including the CLIP score, the comparison of different vectorization methods, and the running time. The experimental results for more applications are also provided. The limitations and the ethical concerns are investigated in the end.

\subsection{Pilot Study Details}

We collect the input images from the internet, and ask 5 artists to give the prompts for these images according to their own tastes. In this way, we obtain 20 input image-prompt pairs for the domain agnostic comparison, and another 20 inputs for the domain-specific comparison. The generated images from different methods are displayed to a total of 38 users. The users are asked to compare the concerned methods from two aspects. The first aspect is the semantic correctness, i.e., how well is the generated image matched with the text prompt. The second aspect is the image quality, i.e., the visual quality of the output image itself. For each input, the users are asked to rank different methods from the best to the worst. The score of each method is determined by its ranking. For the comparison of $K$ methods, the method which is ranked $i$-th is assigned $(K-i+1)$ point(s).

\begin{figure*}[t]
\centering
\includegraphics[width=0.7\linewidth]{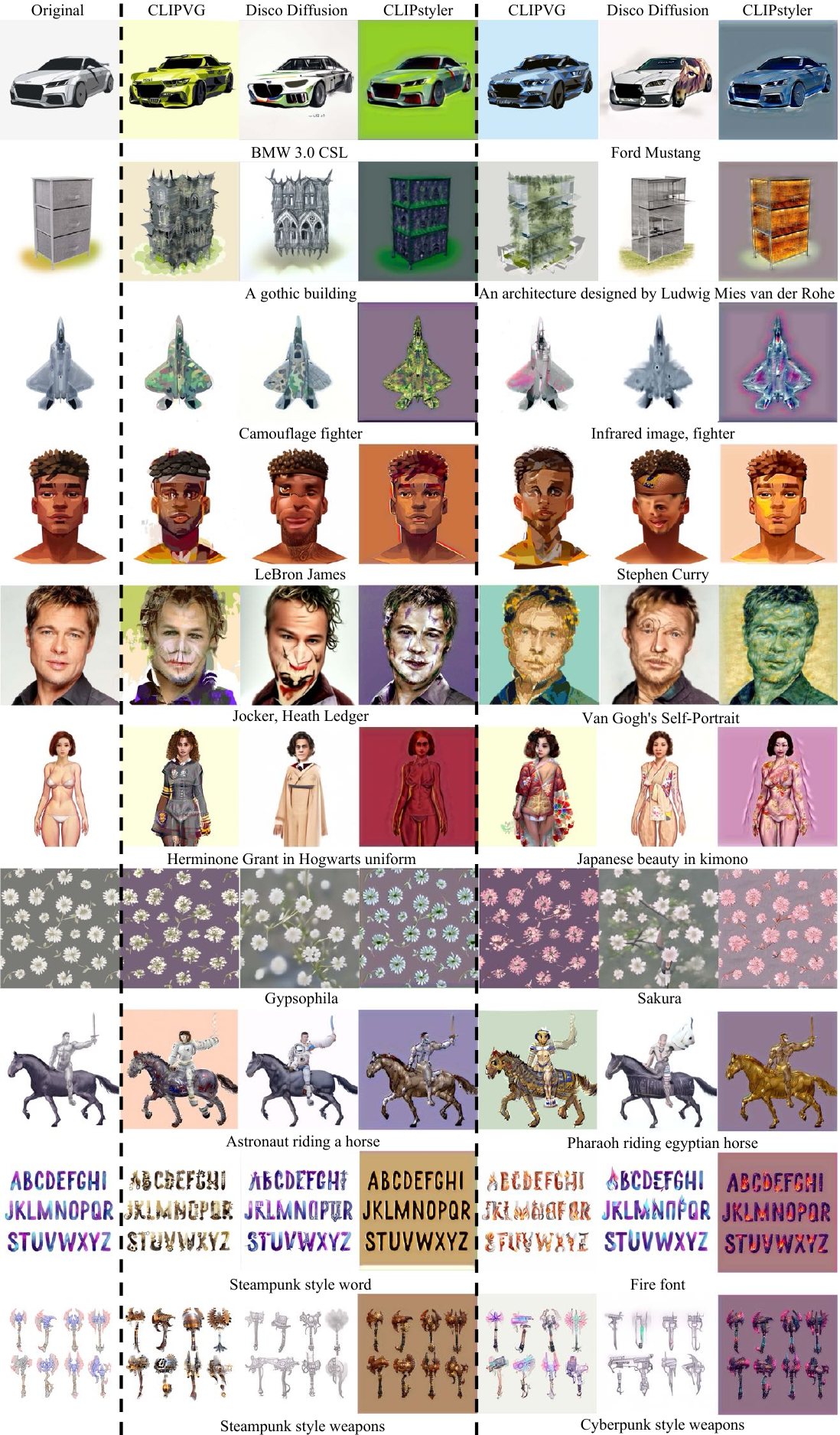} 
\caption{Domain-agnostic comparison for the pilot study.}
\label{fig:pilot_domain_agnostic}
\end{figure*}

\begin{figure*}[t]
\centering
\includegraphics[width=0.95\linewidth]{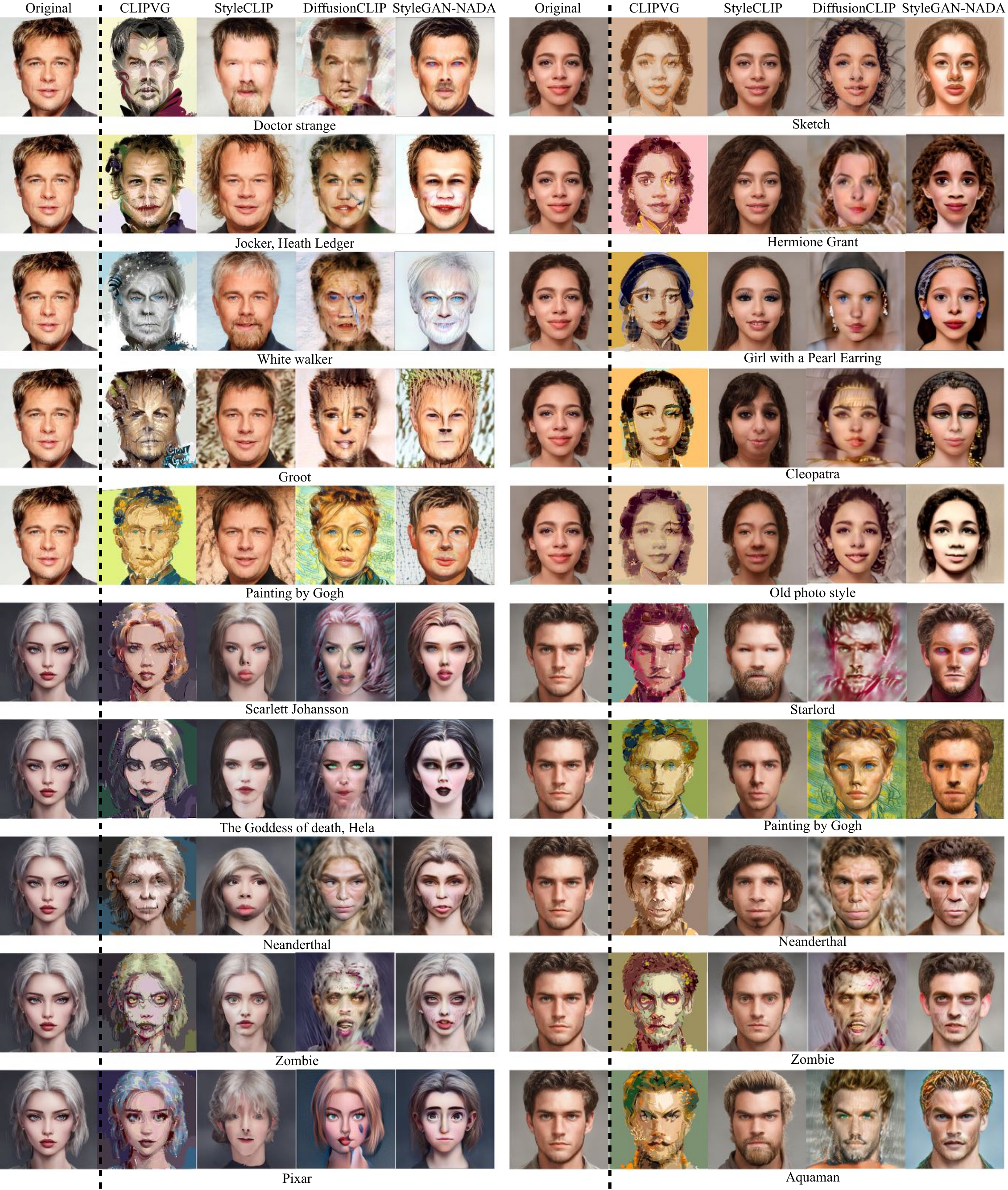} 
\caption{Domain-specific comparison for the pilot study.}
\label{fig:pilot_domain_specific}
\end{figure*}

For the domain-agnostic comparison, we collect 10 input images of different categories, e.g., human body, car, weapon, etc. Two prompts are assigned to each input image.  We use 4 input images of human face for the domain-specific comparison, since human face is the most well-studied target for the domain-specific generative models. Each input image is tested with 5 text prompts in the domain-specific case. All the images and prompts for the domain-agnostic and domain-specific comparisons are shown in Figure~\ref{fig:pilot_domain_agnostic} and Figure~\ref{fig:pilot_domain_specific} respectively.

\clearpage

\clearpage

\subsection{More Evaluations}
\textbf{CLIP Score} Besides the pilot study, we also use CLIP score \cite{glide, diffusionclip, clipstyler} as an additional evaluation metric for the accuracy of semantic manipulation. The CLIP score is defined as the cosine similarity between the CLIP embedding vectors of the text prompt and the output image, i.e., 
\begin{equation}
\label{eqn:clip_score}
\begin{array}{r}
CS=\frac{E_I(I_{out}) \cdot E_T(t_{pr})}{|E_I(I_{out}) || E_T(t_{pr})|},
\end{array}
\end{equation}
where $t_{pr}$ and $I_{out}$ are the input text prompt and output image respectively. $E_T$ and $E_I$ are the text and image encoders of the CLIP model.

\begin{figure}[b]
     \centering
     \begin{subfigure}[b]{0.23\textwidth}
         \centering
         \includegraphics[width=\textwidth]{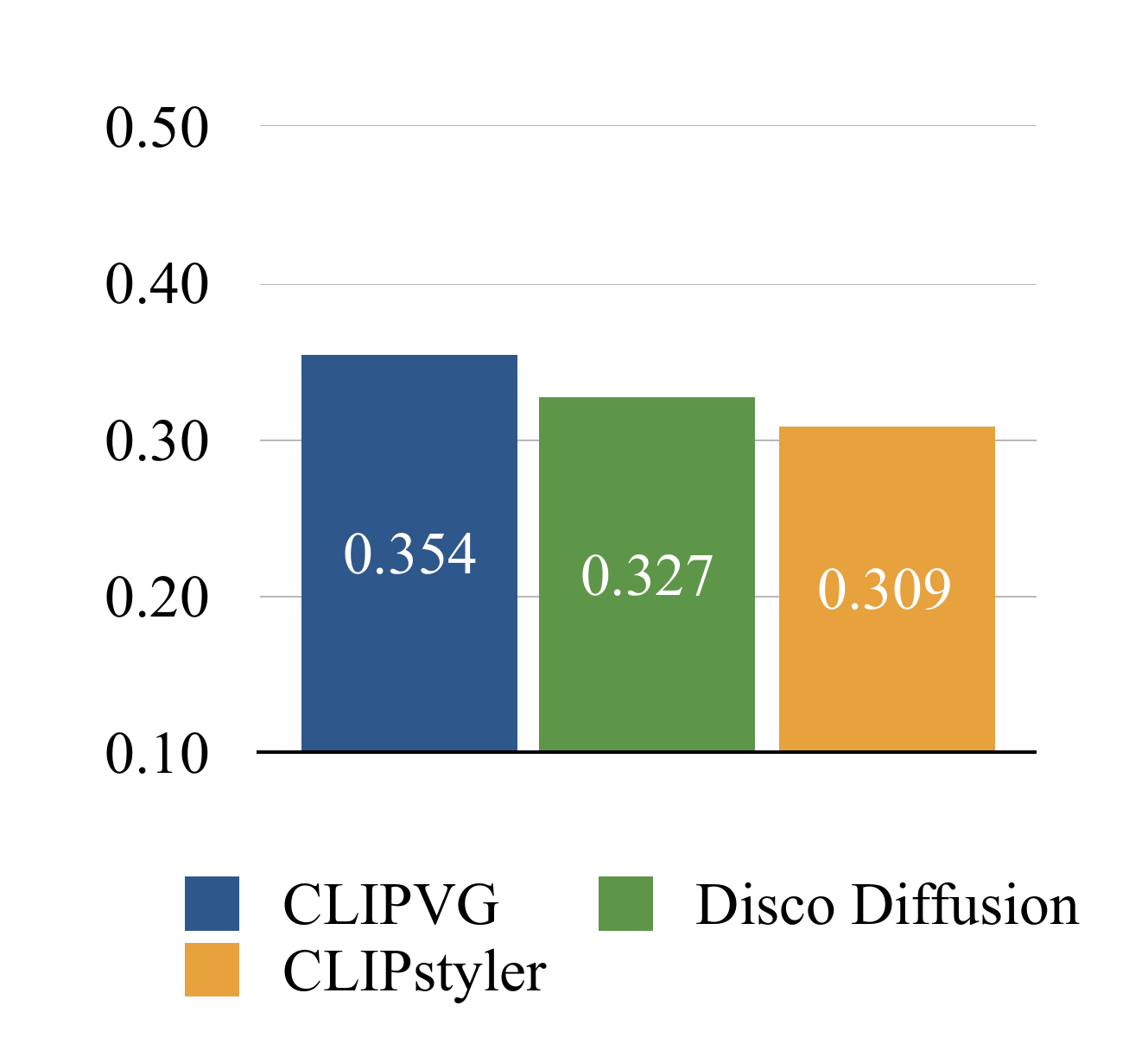}
         \caption{Domain-agnostic.}
         \label{fig:clip_score_agnostic}
    \end{subfigure}
    \hfill
     \begin{subfigure}[b]{0.23\textwidth}
         \centering
         \includegraphics[width=\textwidth]{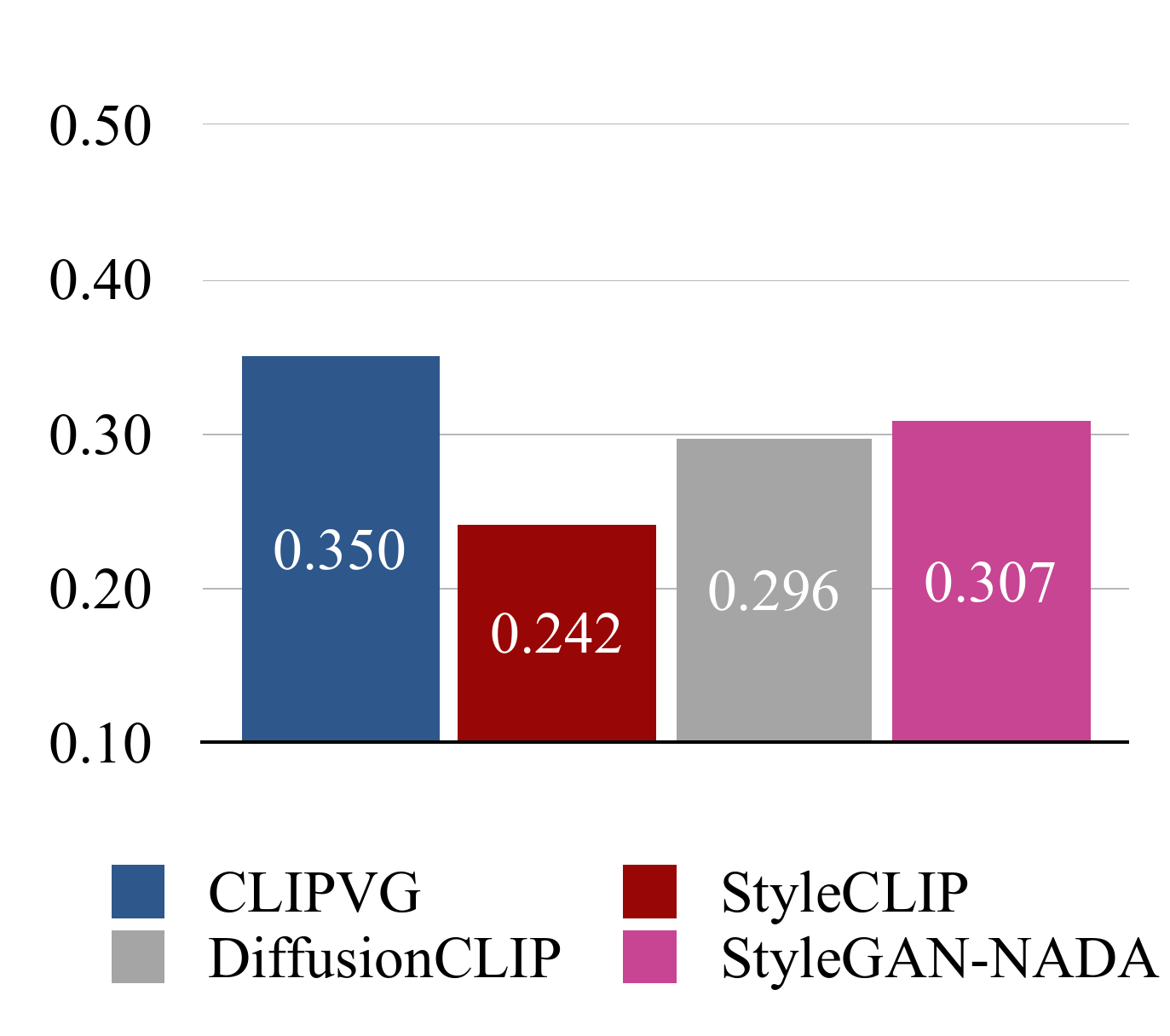}
         \caption{Domain-specific.}
         \label{fig:clip_score_specific}
    \end{subfigure}
\caption{The average CLIP score.} 
\label{fig:clip_score}
\end{figure}

We use the same set of images for the CLIP score and the pilot study. The trend of CLIP score in Figure~\ref{fig:clip_score} is generally consistent with the semantic correctness of the pilot study. CLIPVG achieves the highest average CLIP score for both domain-agnostic and domain-specific comparisons, showing the strong performance for different scenarios. Notice that the CLIP scores of the domain-agnostic and domain-specific scenarios are not directly comparable, since the input images and prompts are different.

\begin{figure}[t]
     \centering
     \begin{subfigure}{\linewidth}
         \centering
         \includegraphics[width=\linewidth]{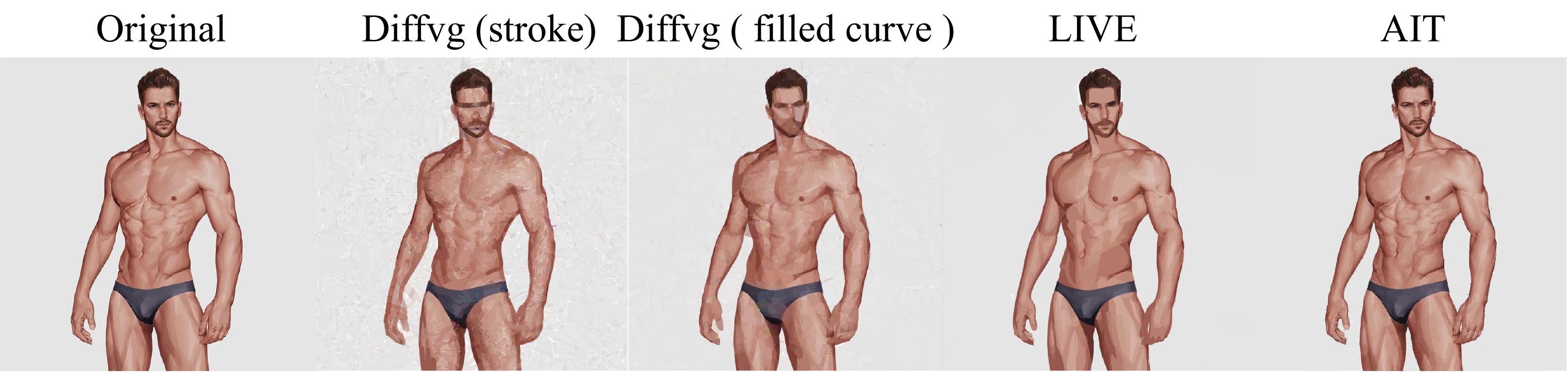}
         \caption{The initialized vector graphics.}
         \label{fig:vectorize_init}
    \end{subfigure}
    \hfill
     \begin{subfigure}{0.8\linewidth}
         \centering
         \includegraphics[width=\linewidth]{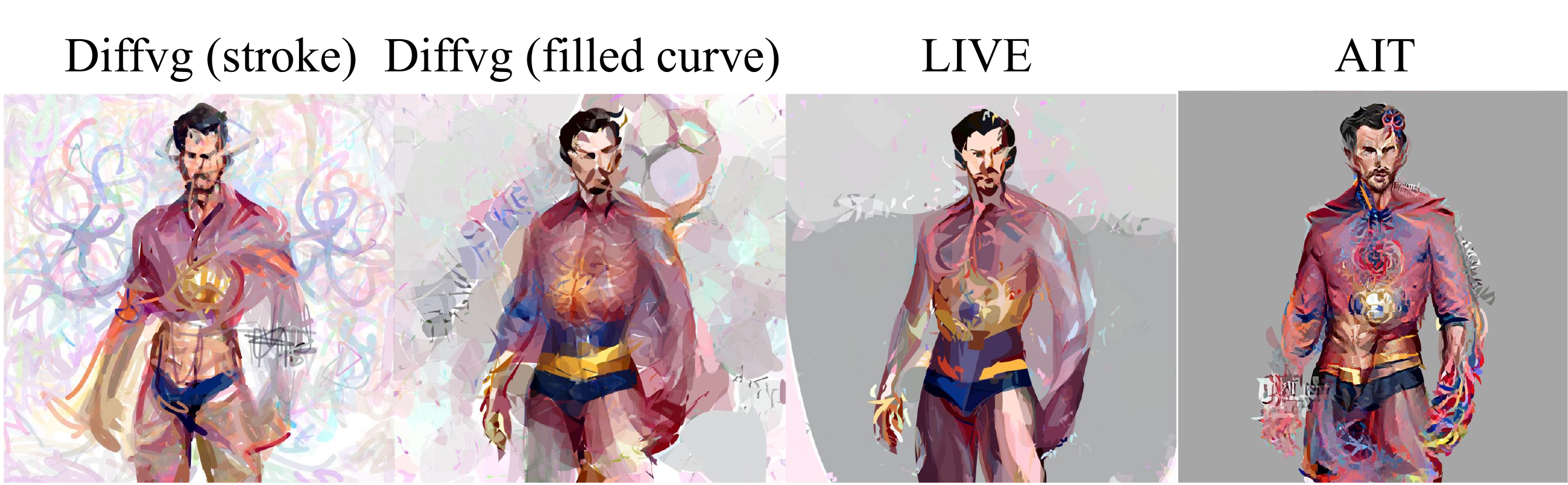}
         \caption{The manipulation results.}
         \label{fig:vectorize_manipulate}
    \end{subfigure}
\caption{Image vectorization and manipulation results with different vectorization tools.}
\label{fig:vectorize}
\end{figure}

\textbf{Vectorization Methods} CLIPVG depends on the existing methods for vectorization. We can either use an open source tool like Diffvg \cite{diffvg} or LIVE \cite{LIVE}, or use a commercial solution like AIT \cite{AdobeImageTrace}. We compare several different vectorization methods in Figure~\ref{fig:vectorize}.
The Diffvg (stoke) and Diffvg (filled curve) methods optimize a set of randomly initialized strokes or filled curves to fit the original image. LIVE gradually adds more filled curves to the canvas in a coarse-to-fine manner. AIT \cite{AdobeImageTrace} is the image tracing tool provided by Adobe Illustrator. We select AIT as our default vectorization tool, since it  gives the most accurate reconstruction result as shown in Figure~\ref{fig:vectorize_init}. The image manipulation results of CLIPVG combined with different vectorization methods are shown in Figure~\ref{fig:vectorize_manipulate}.

\textbf{Running Time.} Our ROI CLIP loss is calculated independently for each different ROI, so the running time increases with more ROI prompts. It typically takes about 0.45 seconds per iteration on a NVIDIA Ampere A30 GPU in the single ROI case, i.e., 67.5 seconds for the full optimization process of 150 iterations. It takes about 0.71 seconds and 0.96 seconds per iteration for the dual-ROI and triple-ROI cases.

Admittedly, CLIPVG runs much slower than some forward-only methods (e.g., DiffusionCLIP \cite{diffusionclip}) which typically only take a few seconds per image, but we argue that these methods require a long training process so the speed comparisons are not fair. CLIPVG is not a low-efficient method compared with other optimization-based baselines. For instance, the typical running times of CLIPstyler \cite{clipstyler} and Disco Diffusion \cite{DiscoDiffusion} are around 40 seconds and 10 minutes per image, respectively.

\begin{figure}[htb]
\centering
\includegraphics[width=0.40\textwidth]{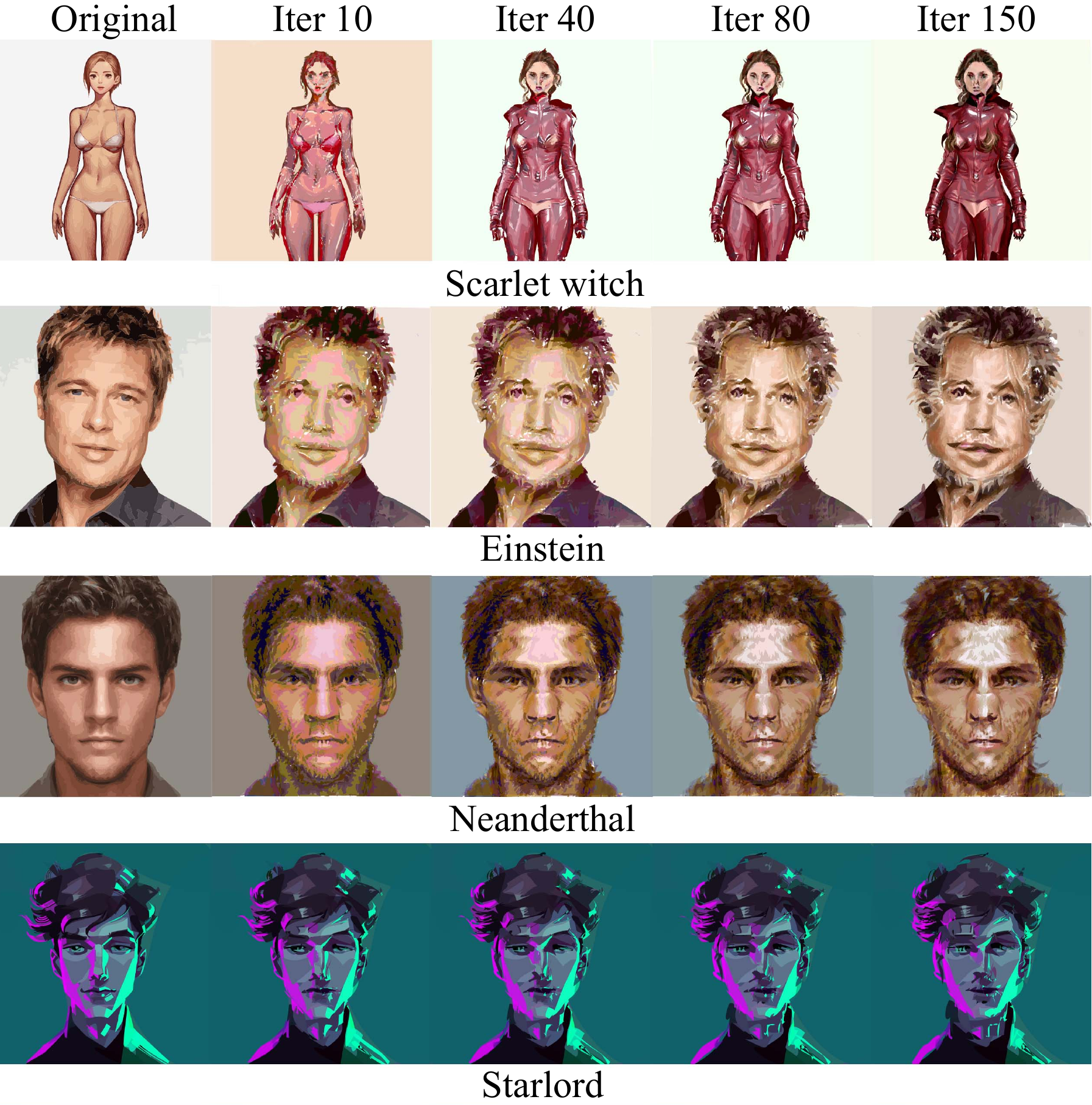}
\caption{Intermediate results during the optimization.}
\label{fig:intermediate_results}
\end{figure}

The overall time overhead of CLIPVG linearly depends on the number of iterations. Therefore, the direct way to speed up CLIPVG is to adaptively choose the iteration number based on CLIP loss or other quality metrics. For conveniences, we set a fixed number (i.e., 150 iterations) for all inputs in the experiments.  Some examples of the intermediates results during the iterative optimization process are shown in Figure~\ref{fig:intermediate_results}. It can be seen that 150 is a rather conservative setting and there is considerable room for reduction.

\subsection{More Applications}

\textbf{Content Loss Based Editing.}  We have developed various fine-grained control methods in the main paper. Here we introduce another simple but effective control method, i.e., applying an additional content loss to constrain the similarity between the output and the input images. The content loss is usually adopted for the applications which require only minor change to the original image. The content loss is defined as $L_{con}(I_{init}, R(\Theta))$, where $L_{con}$ is the content loss function which measures the difference between the initial image $I_{init}$ and the rasterized output image $R(\Theta)$.

\begin{figure}[htb]
\centering
\includegraphics[width=0.4\textwidth]{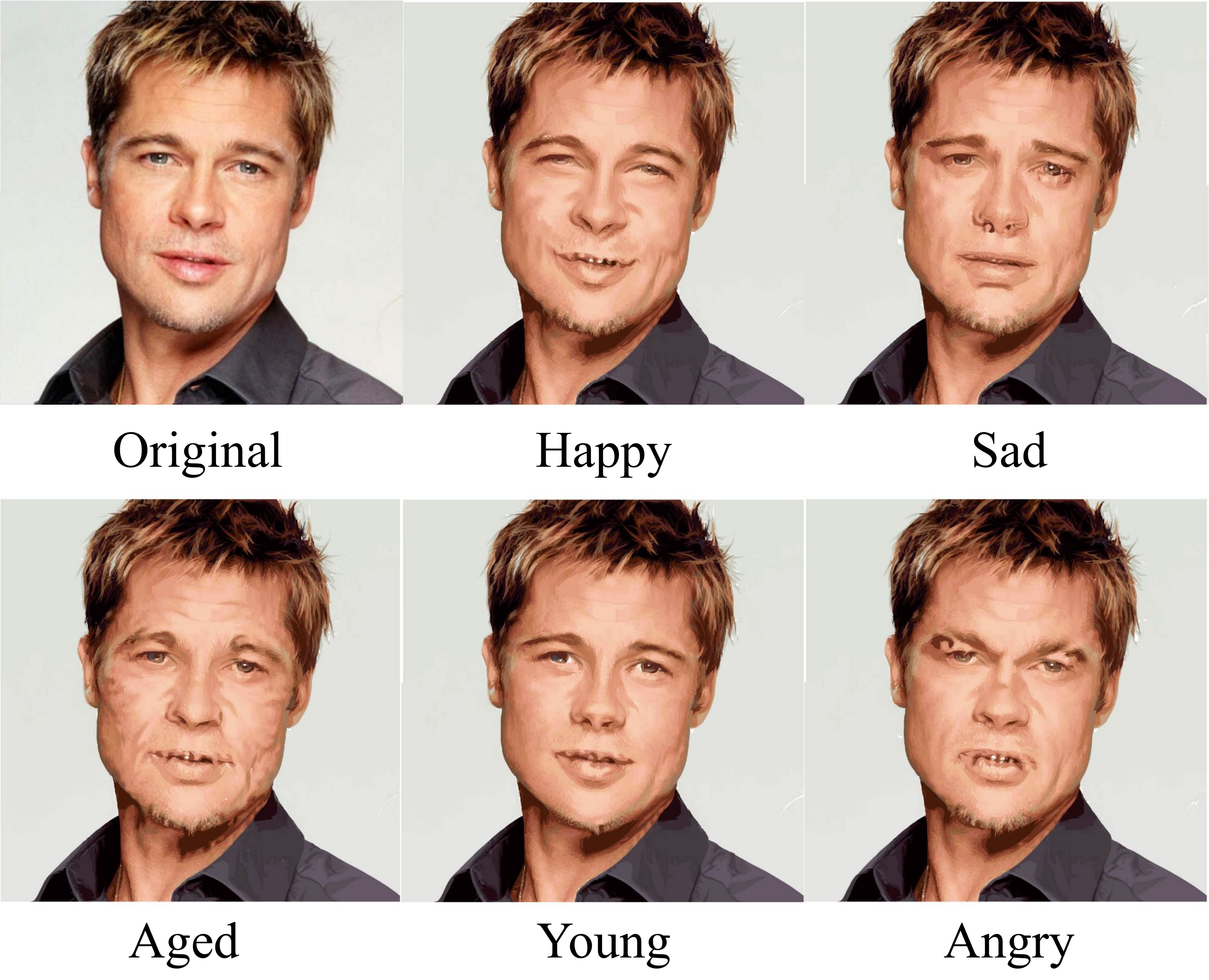} 
\caption{Face attribute editing with the content loss. }
\label{fig:face_attribute}
\end{figure}

The typical content loss functions include L1 or L2 distance, LPIPS \cite{LPIPS}, face identity loss \cite{arcface}, etc. We just consider the L2 loss for simplicity. Since a vector element is corresponding to multiple pixels in the rasterized image, a pixel-level constraint such as L2 is expected to be robust enough. We apply content loss for the application of face attribute editing as shown in Figure~\ref{fig:face_attribute}. The weight for the L2 content loss is set to be 1e6 in this example.

\begin{figure}[htb]
\centering
\includegraphics[width=0.40\textwidth]{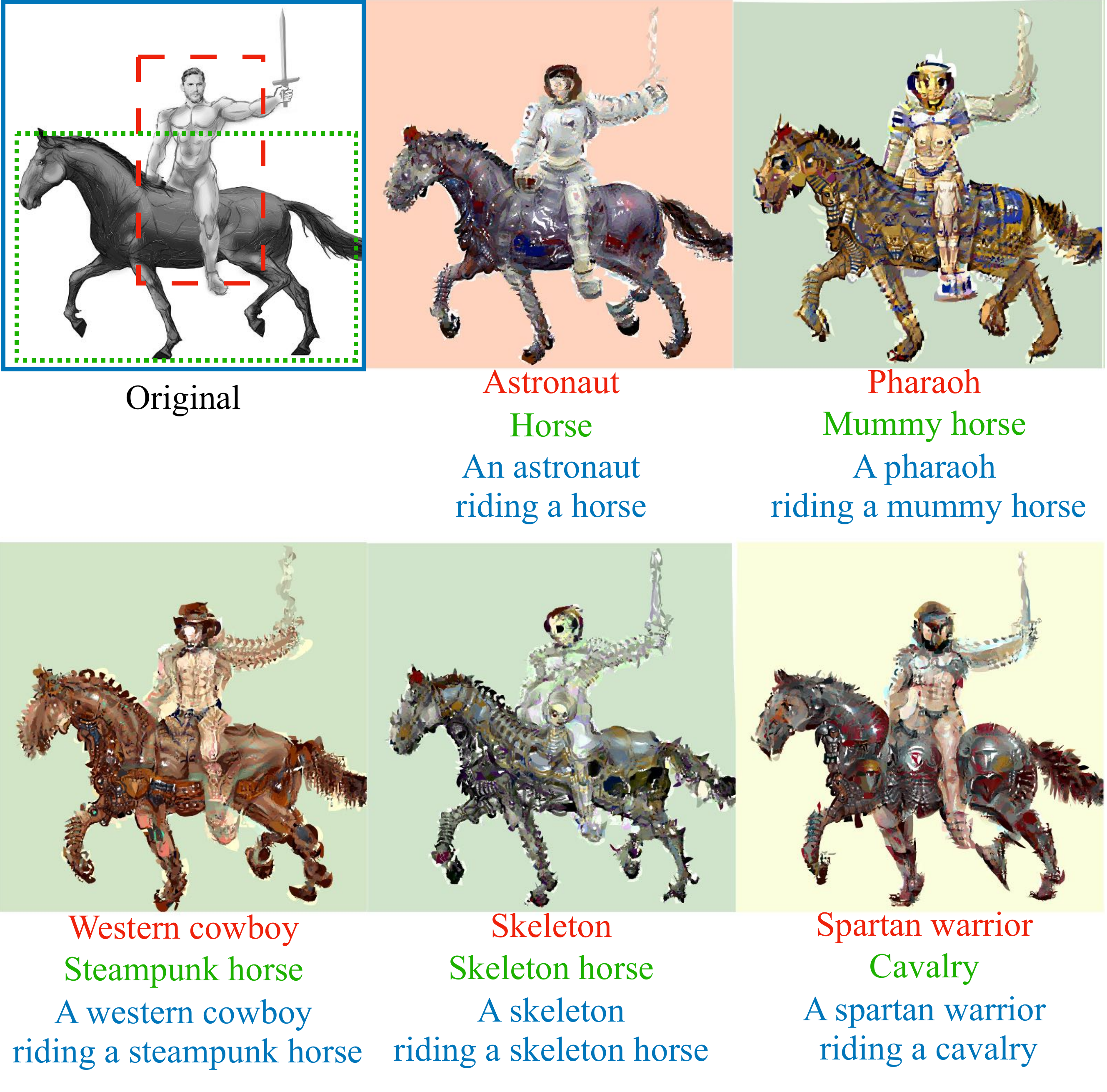} 
\caption{ More results for the ROI prompts. }
\label{fig:roi_prompts_supl}
\end{figure}

\begin{figure}[htb]
\centering
\includegraphics[width=0.47\textwidth]{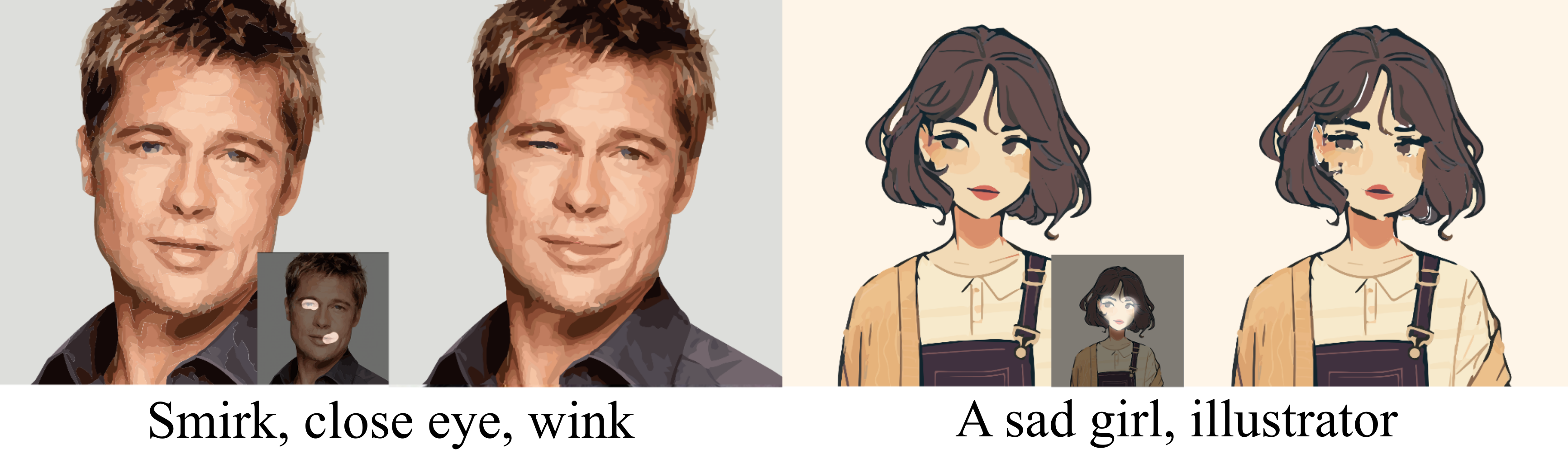} 
\caption{ More subregion editing results. }
\label{fig:subregion_edit_supl}
\end{figure}

\begin{figure*}[htb]
\centering
\includegraphics[width=0.95\textwidth]{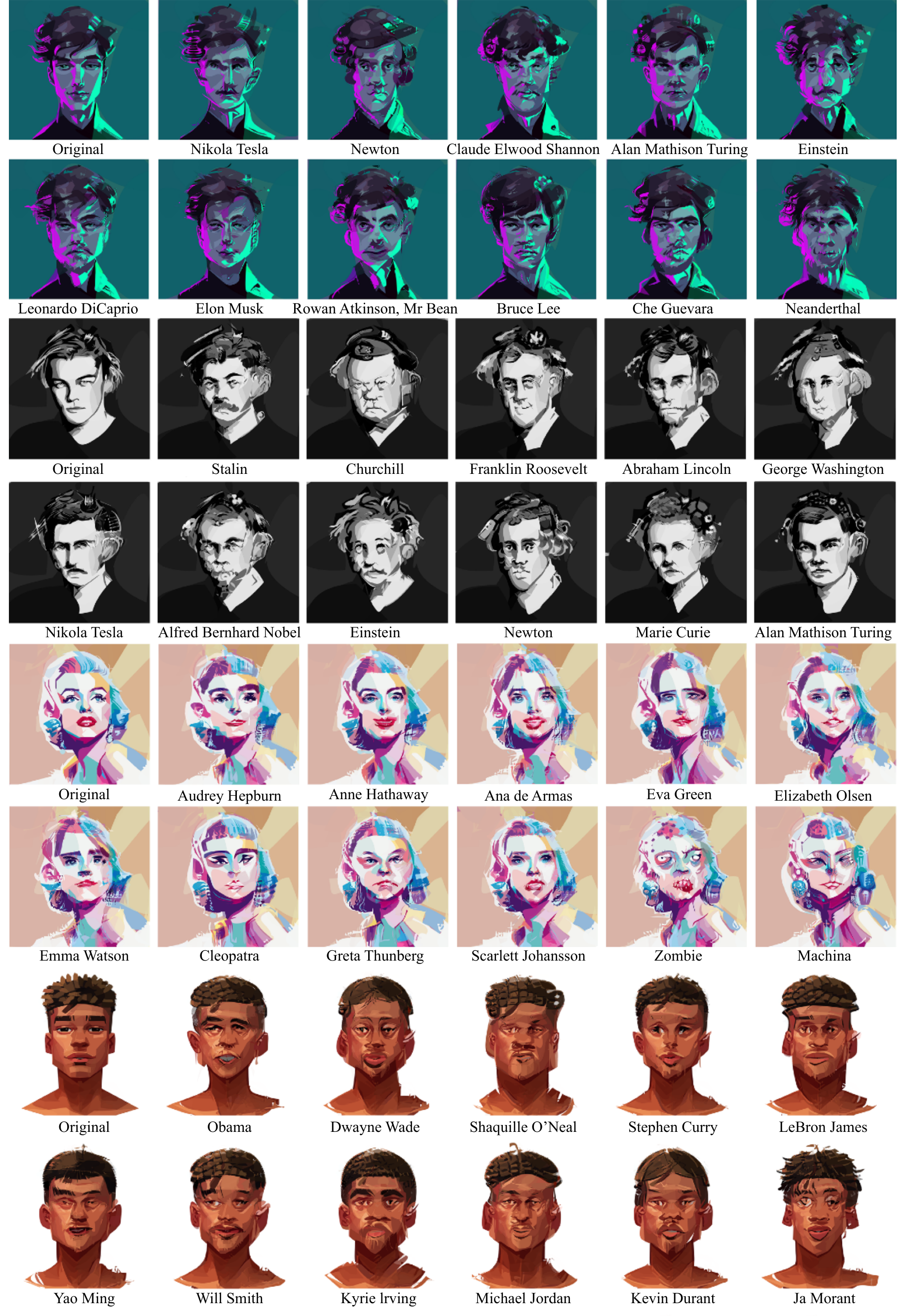} 
\caption{Portrait generation by shape manipulation.}
\label{fig:shape_only_supl}
\end{figure*}

\begin{figure*}[htb]
\centering
\includegraphics[width=0.95\textwidth]{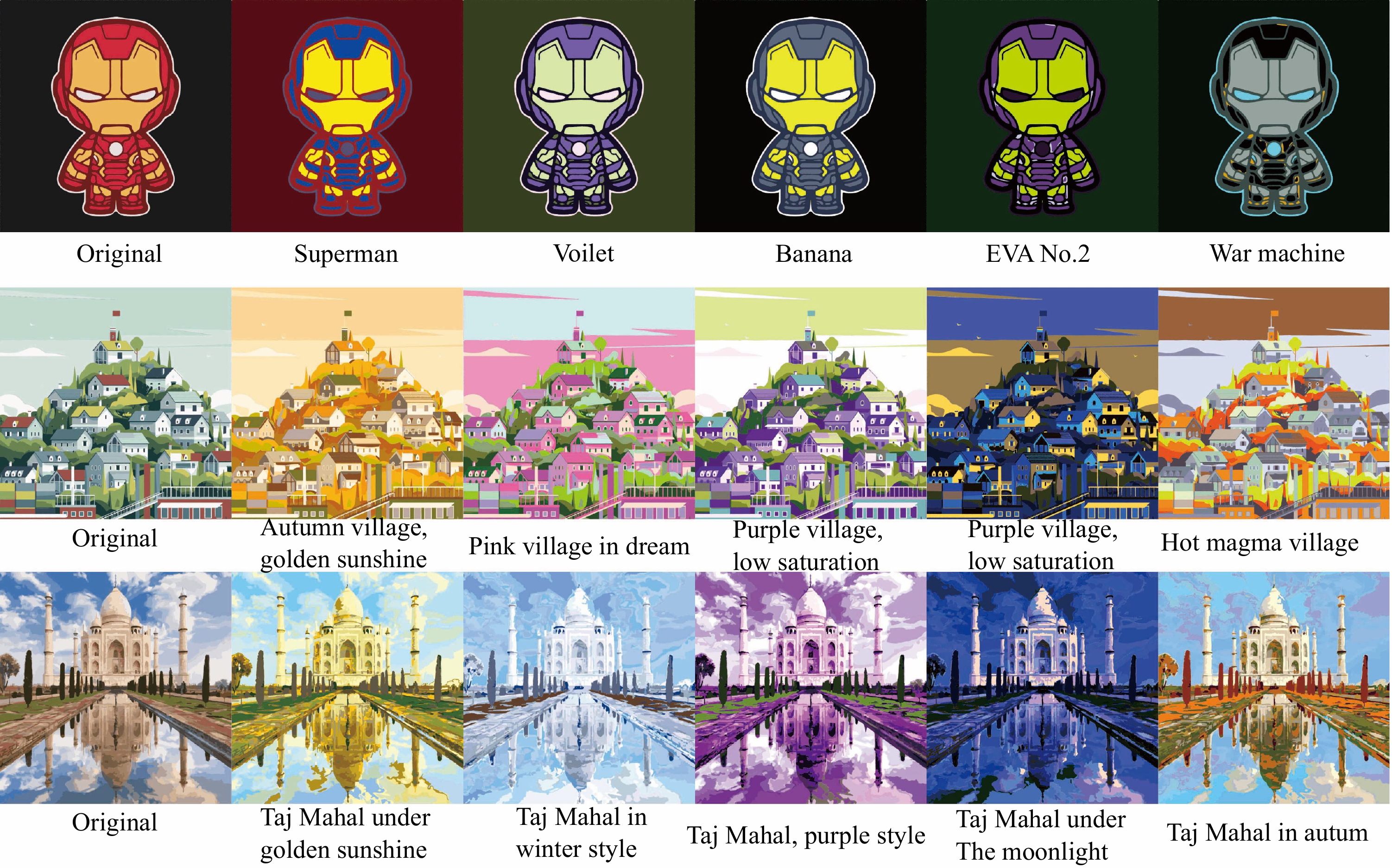} 
\caption{More re-colorization results.}
\label{fig:color_only_supl}
\end{figure*}

\textbf{Fine-Grained Control Experiments.} Extensive experiments have been done for the fine-grained control methods in the main paper. More results for ROI prompts are shown in  Figure~\ref{fig:roi_prompts_supl}. Extra experiments for
subregion editing, shape-only optimization and color-only optimization are shown in Figure \ref{fig:subregion_edit_supl}, \ref{fig:shape_only_supl} and \ref{fig:color_only_supl}.

\begin{figure*}[htb]
\centering
\includegraphics[width=0.95\textwidth]{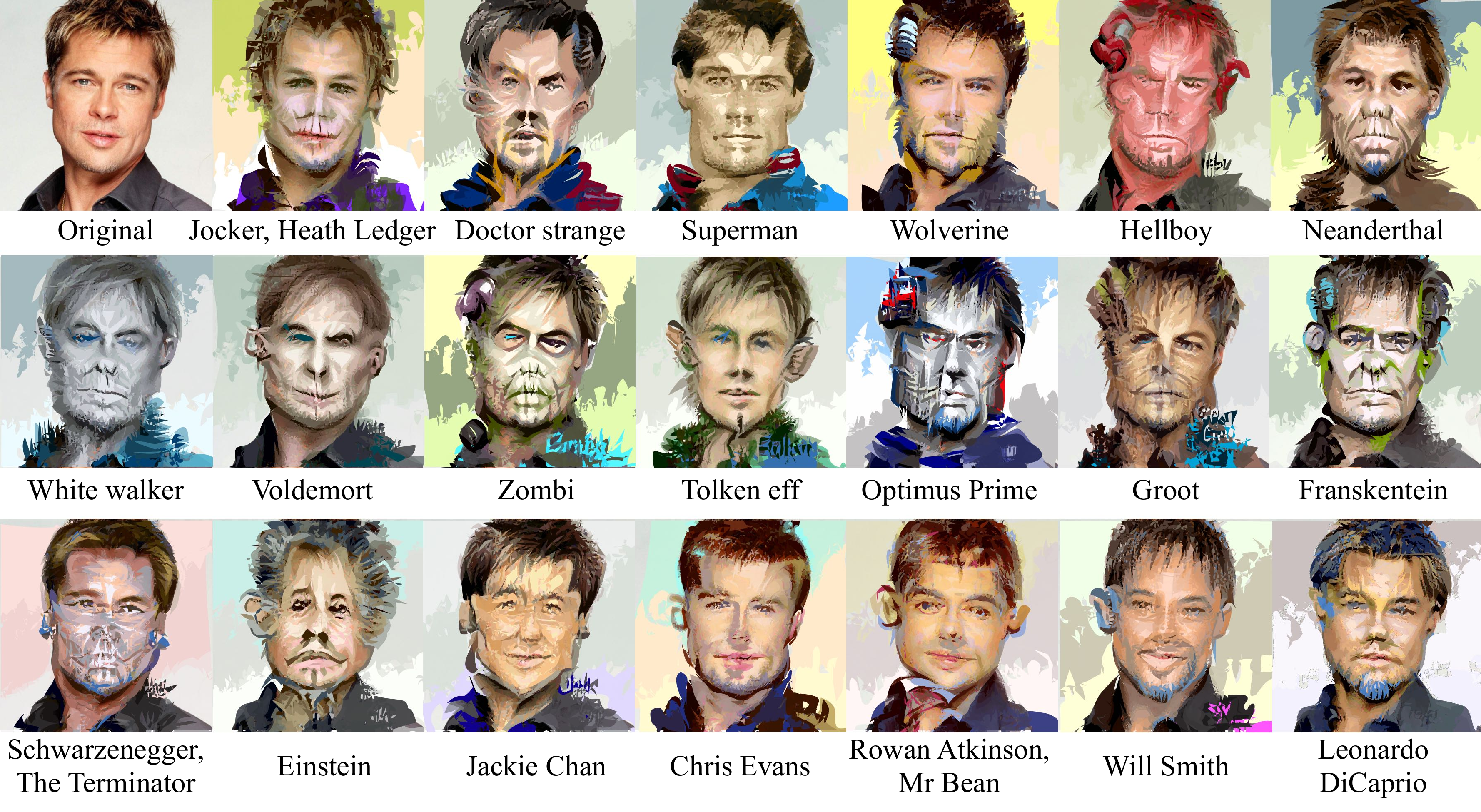} 
\caption{Face Manipulation.}
\label{fig:face}
\end{figure*}

\begin{figure*}[htb]
\centering
\includegraphics[width=0.95\textwidth]{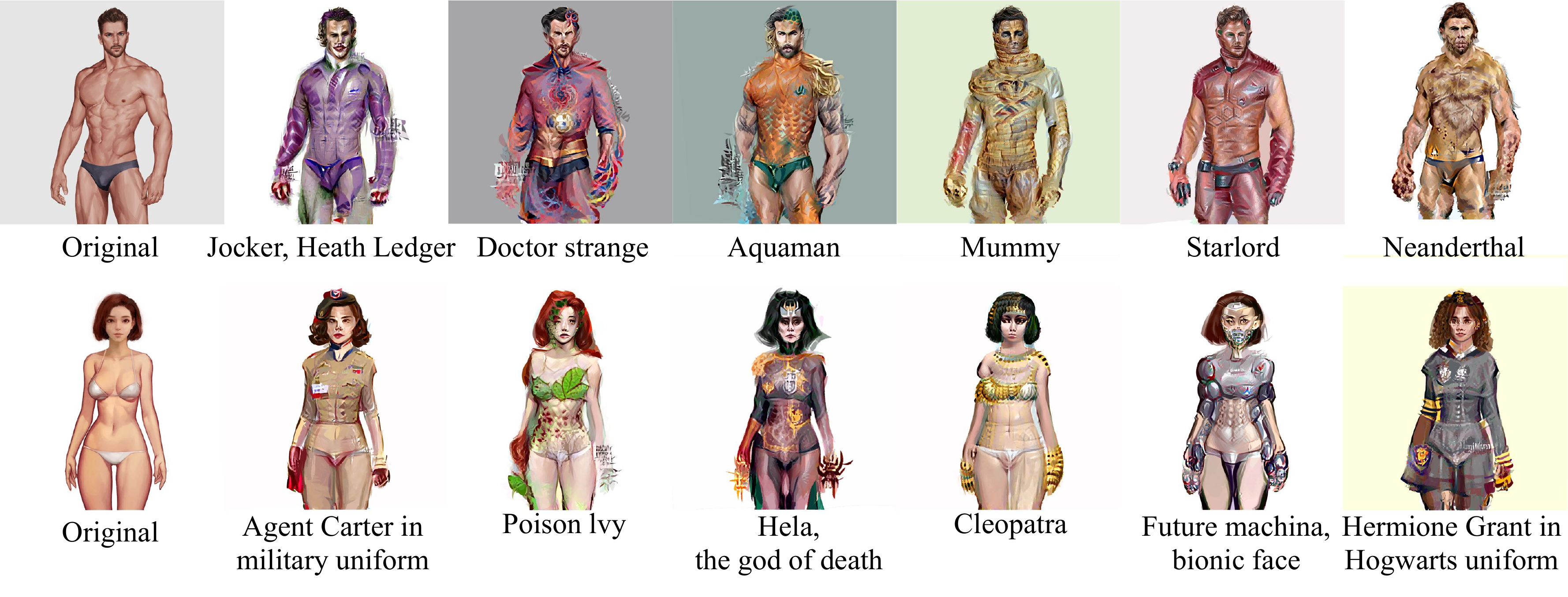} 
\caption{ Character Transformation.}
\label{fig:character}
\end{figure*}

\begin{figure*}[htb]
\centering
\includegraphics[width=0.95\textwidth]{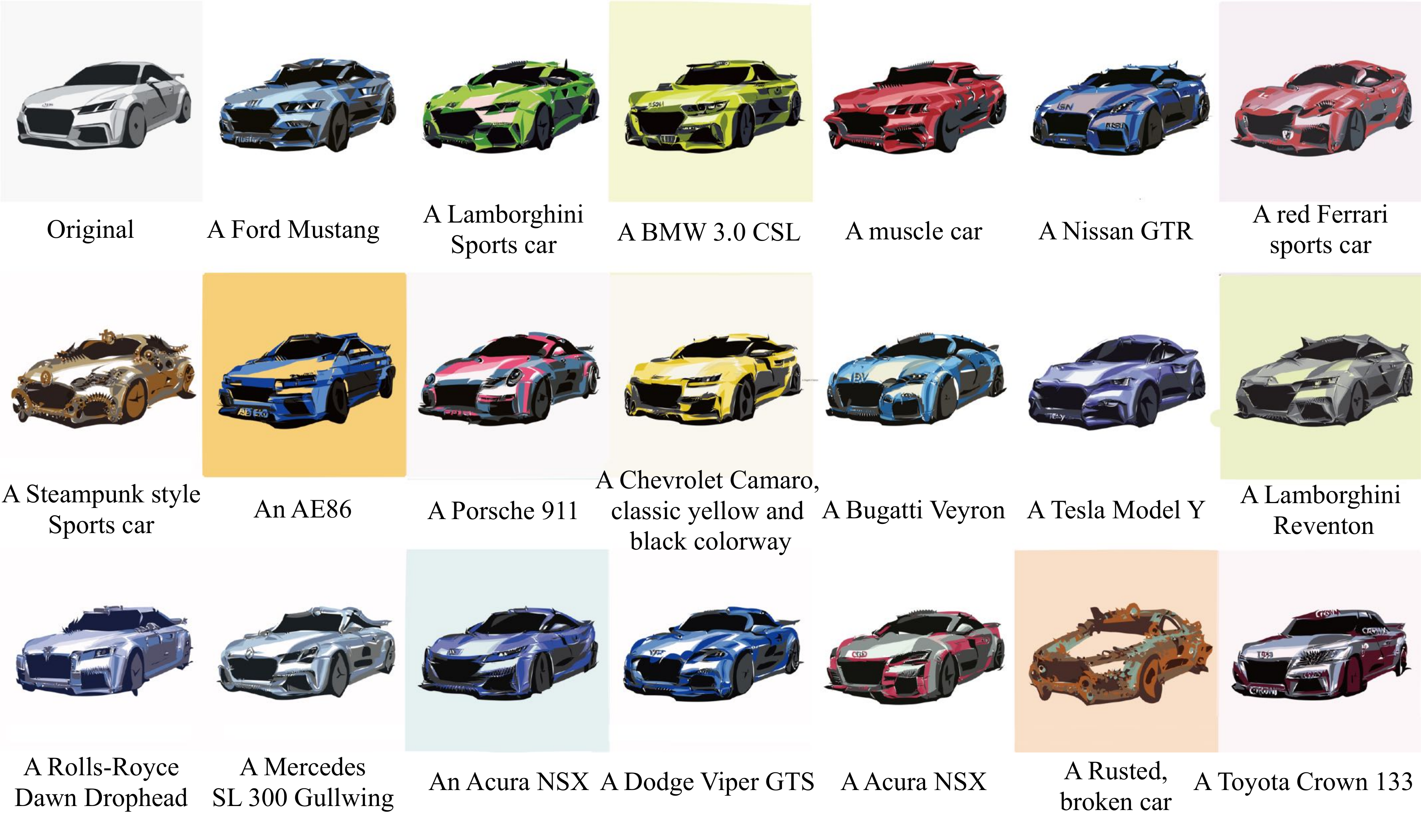} 
\caption{ Car transformation. }
\label{fig:car}
\end{figure*}

\begin{figure*}[htb]
\centering
\includegraphics[width=0.75\textwidth]{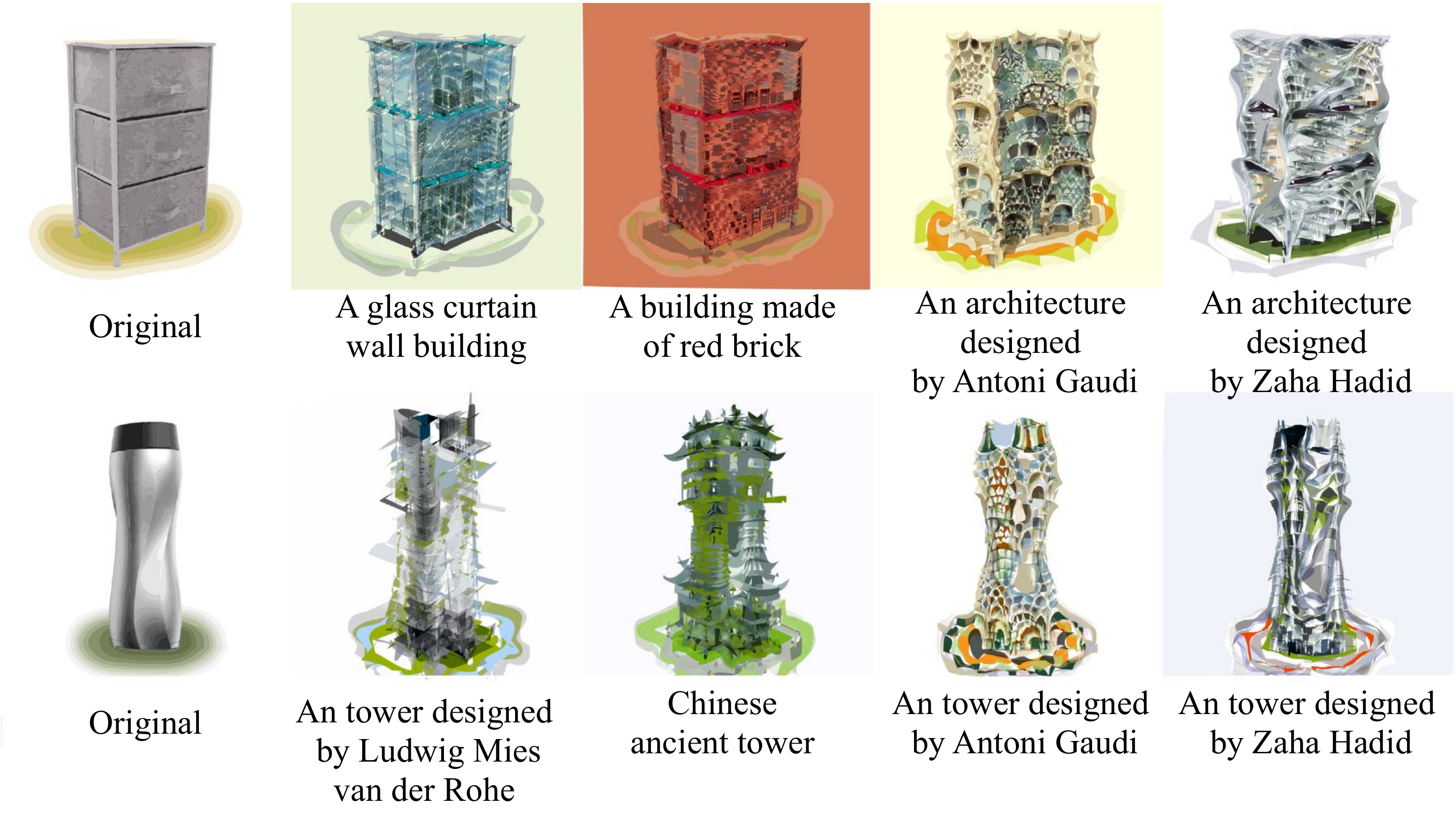} 
\caption{ Architecture Generation.}
\label{fig:architecture}
\end{figure*}

\begin{figure*}[htb]
\centering
\includegraphics[width=0.65\textwidth]{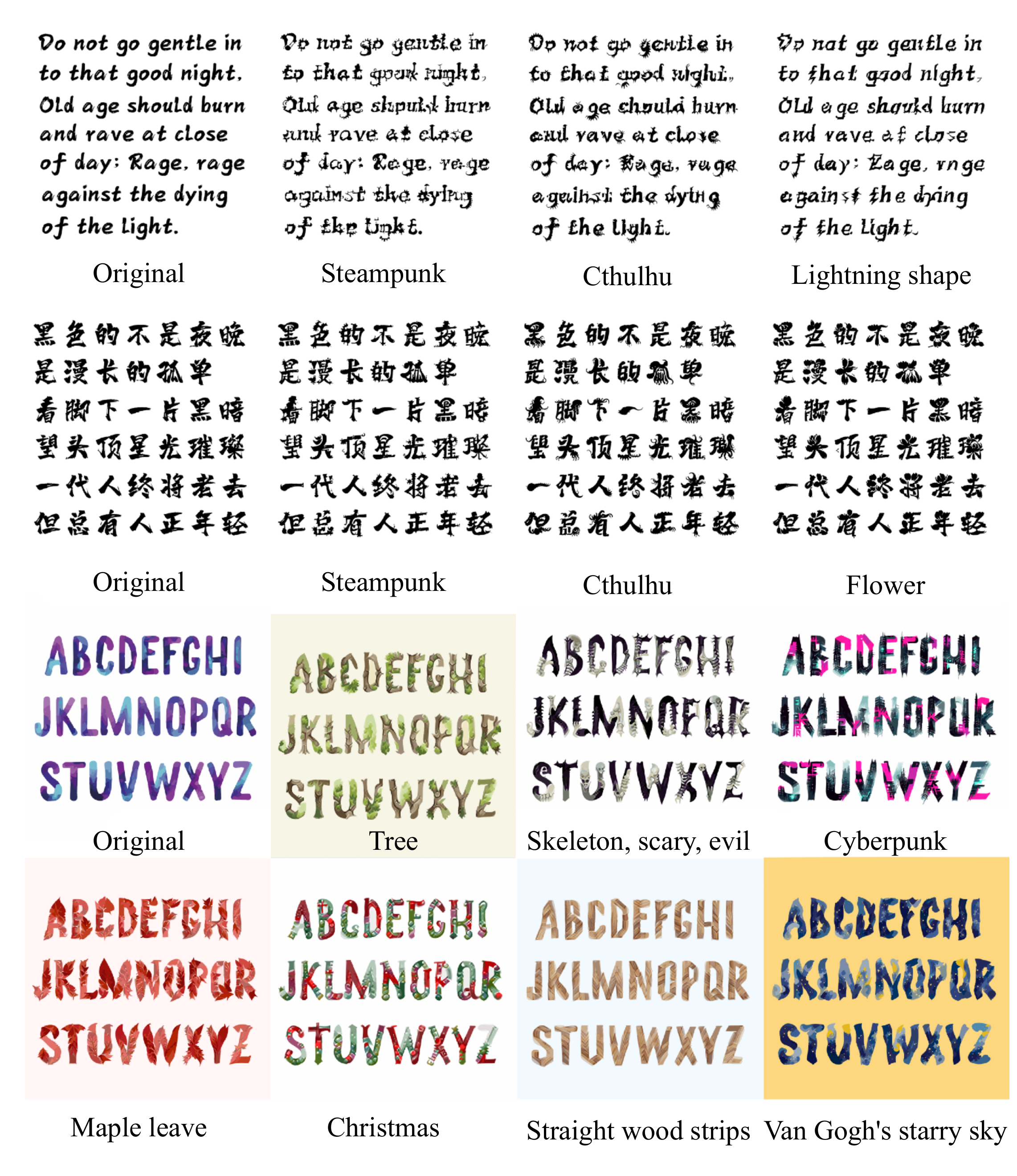} 
\caption{ More font stylization results.}
\label{fig:font}
\end{figure*}

\begin{figure*}[htb]
\centering
\includegraphics[width=0.75\textwidth]{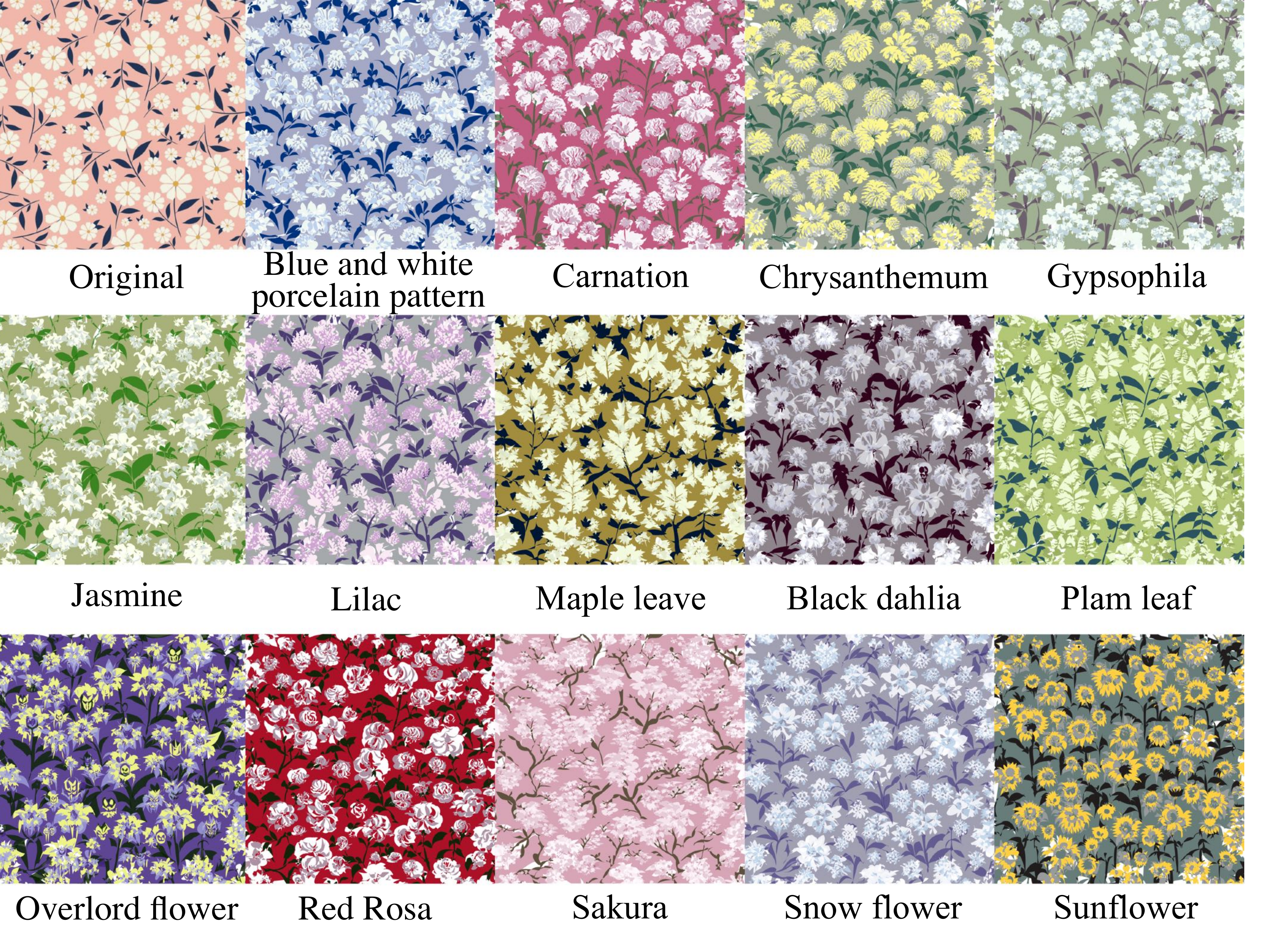} 
\caption{Floral texture design.}
\label{fig:texture}
\end{figure*}

\textbf{Other applications.} In addition to the above experiments, CLIPVG also enables a wide range of other applications. We demonstrate some of these applications including face manipulation, character transformation, car transformation, architecture generation, font stylization and floral texture design in Figure \ref{fig:face}, \ref{fig:character}, \ref{fig:car}, \ref{fig:architecture}, \ref{fig:font} and \ref{fig:texture} respectively.

\subsection{Limitations}
Our method depends on an iterative optimization process which typically takes several minutes for each input image. There are some potential solutions to accelerate the optimization process, e.g., automatic early-stopping according to the CLIP loss, which have not been considered in this work.

Another limitation of our method is the inevitable quantization error of the vectorization process. More specifically, the high frequency information is partially lost during the vectorization. On the other hand, this limitation is likely to be trivial for our task, since the high frequency information of the original image is usually neglected by the image manipulation process.

\subsection{Potential Ethical Concerns}
\textbf{Dataset Bias}. CLIPVG relies on the CLIP model for parsing the text guidance. The pre-trained CLIP model is biased to some extent as described in the CLIP paper \cite{CLIP}. Our method does not rely on any other dataset or pre-trained model besides CLIP.

\textbf{Social impact}. CLIPVG is a domain-agnostic image manipulation method which can be applied to a large number of applications. There is some potential risk of misuse or abuse of our method, such as manipulating an image with malicious text prompts. Different from the raster image based methods, our technique can also be used to create fake or obscene vector graphics. However, as image generation and manipulation tools have become accessible to more and more people, the threat has also become well-aware. Furthermore, the image forensic techniques can be used to distinguish between the original and manipulated images. We strongly suggest that our method should be used cautiously to avoid any negative social impact.

\end{document}